\documentclass[letterpaper]{article} 
\usepackage{aaai2026}  
\usepackage{times}  
\usepackage{helvet}  
\usepackage{courier}  
\usepackage[hyphens]{url}  
\usepackage{graphicx} 
\usepackage{natbib}  
\usepackage{caption} 
\usepackage{algorithm}

\usepackage{algorithmic}
\usepackage{pifont}

\usepackage{newfloat}
\usepackage{listings}
\DeclareCaptionStyle{ruled}{labelfont=normalfont,labelsep=colon,strut=off} 
\floatstyle{ruled}
\newfloat{listing}{tb}{lst}{}
\floatname{listing}{Listing}
\usepackage[export]{adjustbox}
\usepackage{booktabs}   
\usepackage{multirow}   
\usepackage{colortbl}   
\usepackage{siunitx}    
\usepackage{caption}
\usepackage{subcaption}
\usepackage[most]{tcolorbox} 
\usepackage{lipsum}

\usepackage[hyphenbreaks]{breakurl}
\usepackage{enumitem}
\usepackage{graphicx} 

\usepackage{graphicx}
\usepackage{microtype}
\usepackage{graphicx}
\usepackage{booktabs} 
\usepackage{amsmath}
\usepackage{autobreak}
\usepackage{enumitem}
\usepackage[ruled,linesnumbered,algo2e]{algorithm2e}
\usepackage{multirow}
\usepackage{makecell}
\usepackage{xcolor,colortbl}
\usepackage{amsmath}
\usepackage{amssymb}
\usepackage{mathtools}
\usepackage{amsthm}

\usepackage{threeparttable}

\usepackage[capitalize,noabbrev]{cleveref}

\theoremstyle{plain}

\theoremstyle{definition}

\theoremstyle{remark}

\usepackage[textsize=tiny]{todonotes}

\PassOptionsToPackage{numbers, compress}{natbib}

\usepackage[utf8]{inputenc} 
\usepackage[T1]{fontenc}    
\usepackage{url}            
\usepackage{booktabs}       
\usepackage{amsfonts}       
\usepackage{nicefrac}       
\usepackage{microtype}      
\usepackage{xcolor}         

\title{HLPD: Aligning LLMs to \underline{H}uman \underline{L}anguage \underline{P}reference for Machine-Revised Text \underline{D}etection}

\author {
    Fangqi Dai\textsuperscript{\rm 1, \rm 2}\equalcontrib,
    Xingjian Jiang\textsuperscript{\rm 1, \rm 3}\equalcontrib,
    Zizhuang Deng\textsuperscript{\rm 1, \rm 4 \thanks{Corresponding author.}}
}
\affiliations{
    \textsuperscript{\rm 1} School of Cyber Science and Technology, Shandong University, China\\
    \textsuperscript{\rm 2} Xi'an Jiaotong-Liverpool University \\
    \textsuperscript{\rm 3} Chalmers University of Technology\\
    \textsuperscript{\rm 4} State Key Laboratory of Cryptography and Digital Economy Security, Shandong University, China\\

    Daifq7@outlook.com,
    jiangxingjian@mail.sdu.edu.cn,
    dengzz@sdu.edu.cn

}

\begin{document}

\maketitle

\begin{abstract}
To prevent misinformation and social issues arising from trustworthy-looking content generated by LLMs, it is crucial to develop efficient and reliable methods for identifying the source of texts.
Previous approaches have demonstrated exceptional performance in detecting texts fully generated by LLMs. However, these methods struggle when confronting more advanced LLM output or text with adversarial multi-task machine revision, especially in the black-box setting, where the generating model is unknown. 
To address this challenge, grounded in the hypothesis that human writing possesses distinctive stylistic patterns, we propose \emph{Human Language Preference Detection} (HLPD). HLPD employs a reward‐based alignment process, \emph{Human Language Preference Optimization} (HLPO), to shift the scoring model's token distribution toward human‐like writing, making the model more sensitive to human writing, therefore enhancing the identification of machine-revised text. We test HLPD in an adversarial multi‑task evaluation framework that leverages a five‑dimensional prompt generator and multiple advanced LLMs to create diverse revision scenarios.
When detecting texts revised by GPT-series models, HLPD achieves a 15.11\% relative improvement in AUROC over ImBD, surpassing Fast-DetectGPT by 45.56\%. 
When evaluated on texts generated by advanced LLMs, HLPD achieves the highest average AUROC, exceeding ImBD by 5.53\% and Fast-DetectGPT by 34.14\%. Code will be made available at \texttt{https://github.com/dfq2021/HLPD}.
\end{abstract}


\section{Introduction}
As Large Language Models (LLMs) such as GPT-3.5~\cite{openai2022chatgpt} and GPT-4o~\cite{achiam2023gpt} continue to advance in generating convincing texts across diverse fields~\cite{m2022exploring,de2021artificial,fang2023systematic}, public concern regarding their potential misuse has grown~\cite{kelley2021exciting}. 
These models can produce persuasive text that is coherent and contextually appropriate but may be incorrect or misleading in practice.
The cogent responses generated by LLMs are increasingly indistinguishable from human-written content, which poses significant risks~\cite{mckenna2023sources}, such as the spread of disinformation~\cite{bian2024influence} and challenges to social equity~\citep{ferrara2023should}. Moreover, certain high-stakes domains such as legal, medical, need rigorous human authorship where every word is deliberately chosen and fully understood to avoid subtle errors or vulnerabilities. AI revisions, even if minor, can introduce subtle errors or vulnerabilities that undermine the integrity of the text.

To mitigate these risks and maintain trust, it is essential to develop efficient and accurate methods for detecting text that LLMs might produce or modify. 
Existing detection techniques, such as DetectGPT~\cite{mitchell2023detectgpt} and Fast-DetectGPT~\cite{bao2023fast} have performed exceptionally well in identifying text entirely generated by LLMs in a white-box setting~\cite{gehrmann2019gltr}, where the detector evaluates the log probability 
of texts, relying on the observation that machine-generated content generally exhibits higher log-likelihoods compared to human-written texts.
However, such methods struggle when confronted with texts generated by advanced LLMs or machine-revised texts where content initially composed by humans is subsequently polished, rewritten, or expanded by LLMs, particularly under black-box settings where the generating model is unknown.

\begin{figure*}[t]
\setlength{\abovecaptionskip}{0pt}   
\setlength{\belowcaptionskip}{0pt}    
\setlength{\floatsep}{0pt}            
\setlength{\textfloatsep}{0pt}
\setlength{\abovecaptionskip}{5pt}
\setlength{\belowcaptionskip}{0pt}
	\centering
		\centering
		\includegraphics
            [width=0.8
		\textwidth]{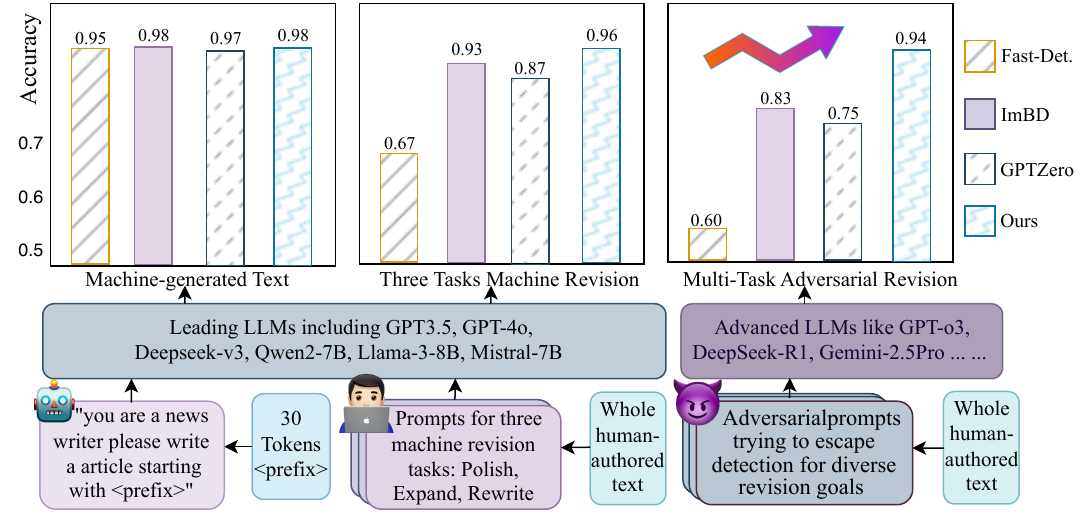}
        \vspace{0pt}
  	     \caption{
    \textbf{Comparison of detecting methods across different scenarios.} Detection accuracy of Fast‑DetectGPT, ImBD, GPTZero, and  HLPD across machine‑generated text (Generate), three tasks revisions (Polish, Expand, Rewrite), and adversarial multi‑task revisions. 
    All methods excel when machine‑style cues are strong (Generate)
    but suffer a sharp drop on revising tasks where machine characteristics are attenuated. ImBD amplifies these cues to outperform Fast‑DetectGPT, yet still degrades on more general situations, whereas HLPD sustains high accuracy across every scenario.
    }
    \vspace{-8pt}
            \label{fig:rewrite}
\end{figure*}
Recent approaches, like ImBD~\cite{tang2024codi}, have improved detection under black-box settings on machine-revised texts by using Style Preference Optimization (SPO) to align the scoring model with machine-generated styles. Nevertheless, as each LLM may have its own unique stylistic characteristics~\cite{reinhart2025llms}, training a scoring model on machine-revised texts from a specific LLM could limit generalization to other models, especially those more advanced ones. Moreover, as illustrated in Figure~\ref{fig:rewrite}, ImBD faces challenges in a more general situation, where texts are revised by state‑of‑the‑art LLMs under diverse, adversarial~\cite{tufts2024practical} prompts.

To overcome these limitations, we first observe that while various LLMs exhibit diverse and unstable linguistic patterns, human writing itself also possesses distinctive stylistic characteristics.~\cite{alafnan2023artificial,guo2023close,pu2023chatgpt}, such as the frequent use of modal and epistemic constructions~\cite{herbold2023large}.  
Leveraging these stylistic differences, we can align the scoring model, adjusting the token distribution of the model to the human language preference rather than machine style, thereby making the scoring model capture human-writing features more efficiently and improving detection accuracy. Existing techniques such as RRHF \cite{yuan2024rrhf} and ORPO \cite{hong2024orpo} have explored the alignment with human preferences, but their primary goal is to enhance the quality of machine-generated text. However, in our work, we adapt the alignment strategy specifically to the scoring task.

Following this principle, we propose a new detection method, \emph{Human language Preference Detection (HLPD)}. First, 
we train the scoring model on paired samples of original human-written text and its corresponding machine-revised version through a reward-based alignment process called \emph{Human Language Preference Optimization (HLPO)}, guiding the model to prioritize features of human-like language and to be more sensitive to human writing. Second, after training, we use the resulting scoring model to compute the metric of \emph{Human Language Preference Conditional Probability Curvature (HLP-CPC)}, capturing the log-probability difference between a candidate text and its perturbed versions, thus facilitating more accurate detection. 

Experimental results demonstrate that HLPD achieves robust performance not only on multi-task adversarial machine revisions, but also excels in detecting texts fully generated by advanced, state-of-the-art LLMs under black-box conditions. 
In single task revision detection on the GPT series, HLPD achieves a 15.11$\%$ relative improvement in AUROC over ImBD, surpasses the logit-based method Fast-DetectGPT by 45.56$\%$ and outperforms the supervised method RoBERTa-large by 38.40$\%$. 
Meanwhile, when detecting adversarial multi-task Revisions, HLPD attains the highest average AUROC, outperforming ImBD by 13.52$\%$ and GPTZero by 24.54$\%$. 
Notably, HLPD also delivers a 5.53$\%$ improvement over ImBD on texts generated by advanced LLMs, showing its effectiveness.




\textbf{Contribution}: 
\ding{182} We propose a novel strategy, HLPD, aligning the scoring model to human language style, significantly enhancing detection accuracy across fully machine-generated and multi-task revised texts under black-box settings. 
Furthermore, we apply the trained scoring model in an adaptive attack on GPTZero to highlight its broader potential.
\ding{183} We design and implement a comprehensive adversarial multi-task evaluation framework that better simulates real user scenarios. Employing a five‐dimensional adversarial prompt generator together with multiple state‐of‐the‐art LLMs to systematically construct diverse machine‐revision scenarios for assessing the effectiveness and robustness of the detection method, i.e., HLPD. 
\ding{184} HLPD achieves significant accuracy improvements across various machine revision tasks and multiple languages in the black-box setting. Showcasing its robustness and adaptability to various scenarios and different advanced models.

\begin{figure*}[t]
\setlength{\abovecaptionskip}{-3pt}   
\setlength{\belowcaptionskip}{0pt}    
\setlength{\floatsep}{0pt}            
\setlength{\textfloatsep}{0pt}
    \centering
   \includegraphics
            [width=1
		\textwidth]{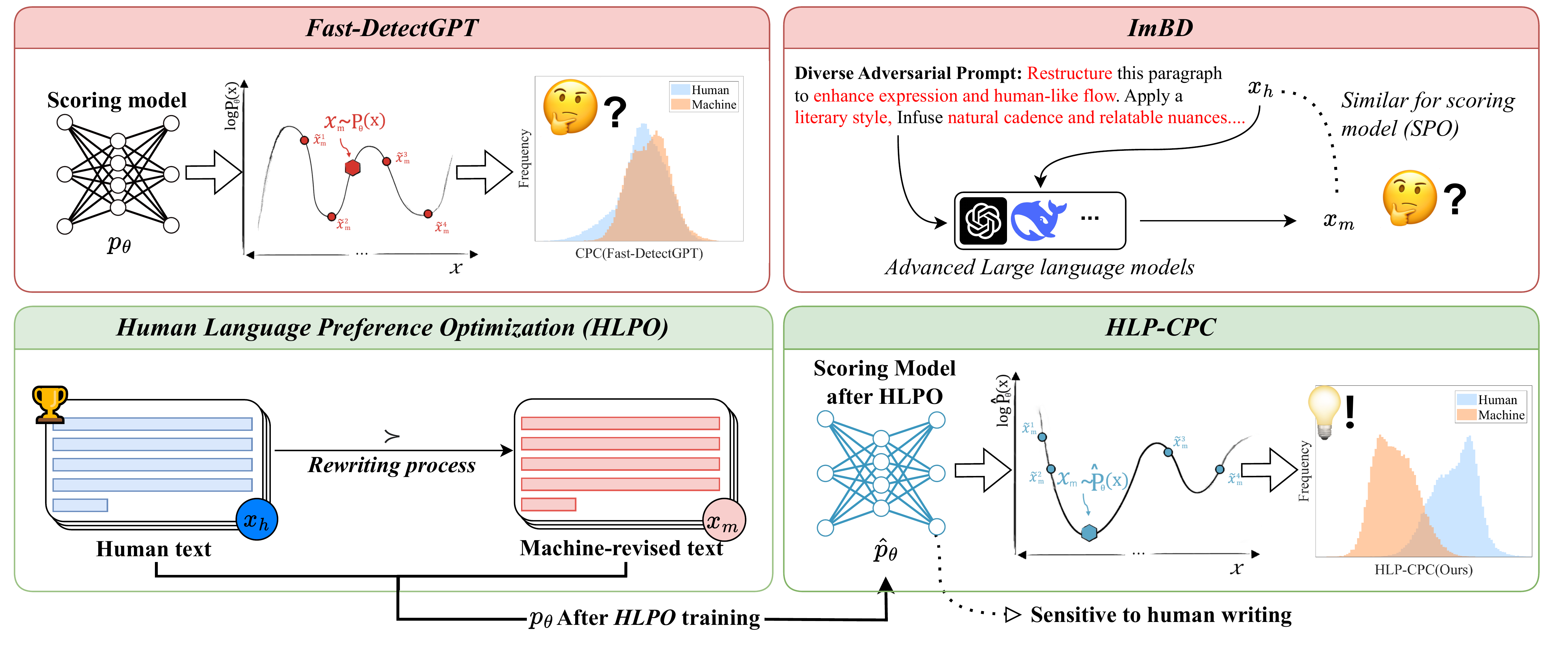}
     \vspace{-5pt}
    \caption{
    \textbf{Challenges of Fast-DetectGPT and ImBD under adversarial revision tasks and overview of HLPD.} 
    \emph{Top row:} On the left, logit-based detectors such as Fast-DetectGPT struggle when texts $x_h$ are only revised, due to weak machine signals. On the right, in ImBD, scoring model after SPO still favors machine-like signals.
     This preference can limit generalization, especially when faced with revisions from advanced models under diverse prompts.
    \emph{Bottom row:} HLPO forms paired human-written and machine-revised texts to train model $p_\theta$, aligning it directly to human writing style. With the trained $\hat{p}_\theta$, HLPD can reliably detect subtle deviations across various scenarios, including minimal revisions and adversarially generated texts by state-of-the-art LLMs, thereby significantly improving robustness in black-box settings.
}
    \label{fig:detect}
     \vspace{-8pt}
\end{figure*}
\vspace{-5pt}
\section{Methodology}
\label{sec:Method}

We present a novel method, \emph{Human Language Preference Detection (HLPD)}, for detecting both machine‑revised and fully generated texts by aligning the scoring model to human writing styles. This alignment improves robustness in the black-box setting, where the specific LLM is unknown, and improves sensitivity to outputs from state-of-the-art LLMs in diverse revision and generation scenarios. For more details of those scenarios, please refer to Section~\ref{subsec:datasets} and Appendix B.

\vspace{-5pt}
\subsection{Foundation}
\label{subsec:foundation}
Style alignment involves fine-tuning a language model to prefer specific stylistic characteristics. In the context of text detection, aligning a scoring model to a particular style, whether machine or human, enables the model to become more sensitive to deviations from that style. This sensitivity widens the gap of log-probabilities difference between the target text \( x \) and its perturbed versions, \( \tilde{x} \). This gap is crucial for accurately identifying texts altered or generated by LLMs.


\vspace{-5pt}
\subsection{Human Language Preference Optimization}
\label{subsec:human_style_preference_optimization}
As illustrated in Figure~\ref{fig:detect}, the essence of this method lies in the pre-training part. Before using the scoring model in calculating Conditional Probability Curvatures (CPC), we need to create preference relations between pairs of texts with identical content: one written by a human ($x_h$) and one revised by a machine ($x_m$). To align the scoring model with human stylistic preferences, we introduce an HLPO process. This process involves training the scoring model to assign higher log probabilities to human-written texts compared to their machine-revised counterparts.

Given a dataset \( \mathcal{D} = \{(x_h^{(i)}, x_m^{(i)})\}_{i=1}^N \), where each pair consists of a human-written text \( x_h^{(i)} \) and its machine-revised version \( x_m^{(i)} \), our goal is to train the scoring model to prefer human style. Specifically, we establish preference relations where the human-written text is preferred over the machine-revised text:
{\setlength{\belowdisplayskip}{3pt} 
\begin{equation}
    x_h^{(i)} \succ x_m^{(i)}, \quad \forall i \in \{1, \dots, N\}.
\end{equation}
We adopt a preference-based optimization framework inspired by the Direct Preference Optimization (DPO) approach~\cite{rafailov2024direct}. The probability of preferring one text over another is modeled using the Bradley-Terry model~\cite{bradley1952rank}:
\begin{equation}
    P(x_h \succ x_m) = \sigma\left( r(x_h) - r(x_m) \right),
\end{equation}
where \( r(x) \) is a reward function representing the model's preference for text \( x \), and \( \sigma \) is the sigmoid function. To make the scoring model prefer the human language style, we need to increase the value of  $r(x_h) - r(x_m)$.

Following DPO, we relate the reward function to the log probabilities assigned by the scoring model \(p_\theta\) we are training and a reference model \( p_{\theta_{\text{ref}}} \), which can simply be the original scoring model before training:

\vspace{-10pt}
\begin{equation}
    r(x) = \beta \left( \log p_\theta(x) - \log p_{\theta_{\text{ref}}}(x) \right),
\end{equation}


where \( \beta \) is a temperature parameter controlling the sharpness of the preference.
At this point, The optimization $\max_{\theta} \mathbb{E}_{(x_h, x_m) \sim \mathcal{D}} \left[ \log \sigma \left( r(x_h) - r(x_m)\right)\right]$ becomes:
\begin{multline}
\max_{\theta}\,\mathbb{E}_{(x_h,x_m)\sim\mathcal{D}}
\bigl[\log\sigma\bigl(\,
\beta(\log p_\theta(x_h)-\log p_\theta(x_m)) \\
\;\;-\;\beta(\log p_{\theta_{\mathrm{ref}}}(x_h)-\log p_{\theta_{\mathrm{ref}}}(x_m))
\bigr)\bigr].
\end{multline}
Since \( p_{\theta_{\text{ref}}} \) is fixed during training, we can simplify the optimization to focus on maximizing the difference \( \log p_\theta(x_h^{(i)}) - \log p_\theta(x_m^{(i)}) \), thus adjusting the scoring model \( p_\theta \) to assign higher probabilities to human-written texts. And to acquire $\max_{\theta} \mathbb{E}_{(x_h, x_m) \sim \mathcal{D}} \left[ \log \sigma \left( r(x_h) - r(x_m) \right) \right]$, equal to acquire $\min_{\theta} -\mathbb{E}_{(x_h, x_m) \sim \mathcal{D}} \left[  r(x_h) - r(x_m)  \right]$. With limiting training data, to directly enlarge the gap of $\log p_{\theta}(x_h) \;-\; \log p_{\theta}(x_m)$, address potential problems like saturation and avoid overfitting in small‑sample preference tuning, based on our experiment, we remove the sigmoid function and introduce an adaptive coefficient \(\beta_t\), yielding the linear contrastive loss:
{\setlength{\belowdisplayskip}{3pt} 
\begin{align}
    \min_{\theta}\;\mathcal{L}(\theta)
    &= -\,\mathbb{E}_{(x_h,x_m)\sim\mathcal{D}}\bigl[\beta_t\;\cdot\;r(x_h,x_m)\bigr].
\end{align}
where
\setlength{\belowdisplayskip}{2pt} 
\begin{multline}
r(x_h,x_m)
= \bigl[\log p_\theta(x_h) - \log p_\theta(x_m) \bigr]\\
\quad -\,\bigl[\log p_{\theta_{\mathrm{ref}}}(x_h) - \log p_{\theta_{\mathrm{ref}}}(x_m)\bigr].
\end{multline}

Here, \(\beta_t\) is adjusted on‑the‑fly via a lightweight variance‑aware scheduler (Dynamic‑\(\beta\)):
\setlength{\belowdisplayskip}{2pt} 
\begin{gather}
    \beta_t \;=\;\mathrm{Dynamic\mbox{-}}\beta\ \bigl(\mathrm{Var}_{\mathrm{window}}[r(x_h,x_m)]\bigr)
\end{gather}

At each step, we compute the variance of the margin \(r(x_h, x_m)\) over a sliding window, increasing \(\beta\) when the variance is low (model confident) and decreasing it when the variance is high (noisy signal). This stabilizes early training, prevents overfitting, and sharpens the margin later, without extra theoretical assumptions. This supervised fine-tuning process ensures that the scoring model \( \hat{p}_\theta \) becomes attuned to the stylistic nuances of human writing, thereby enhancing its ability to detect deviations introduced by machine revisions. We present more detailed comparisons of different loss functions in ablation studies in Section ~\ref{Ablation}.


\vspace{-2pt}
\subsection{Detection via HLP-Conditional Probability Curvature}
\label{subsec:style_cpc}
\vspace{-2pt}
As shown in Figure~\ref{fig:detect}, after aligning the scoring model with human writing styles, we employ a modified probability curvature metric to detect machine-revised text. This metric, termed \emph{Human Language Preference Conditional Probability Curvature (HLP-CPC)}, builds upon the probability curvature introduced in Fast-DetectGPT and adapts it to our human style-aligned scoring model.

In Fast-DetectGPT, the key observation is that machine-generated texts often occupy regions of negative curvature in the log probability landscape, as they tend to have higher log probabilities under the language model compared to human-written texts. However, since our scoring model \( \hat{p}_\theta \) is aligned to prefer human writing styles, the situation is reversed: human-written texts receive higher log probabilities, and machine-revised texts receive lower log probabilities.

To quantify this, given a passage \( x \), the aligned scoring model \( \hat{p}_\theta \), and the perturbation model \( q_\phi \),  we define the conditional probability function as:
\vspace{0pt}
\begin{equation}
    \hat{p}_\theta(\tilde{x}|x) = \prod_{j=1}^n \hat{p}_\theta(\tilde{x}_j | x_{<j}),
\end{equation}
where \( \tilde{x}_j \) is the \( j \)-th token in the perturbed passage \( \tilde{x} \), generated by sampling from the conditional distribution \( \hat{p}_\theta(\tilde{x}_j | x_{<j}) \), and \( x_{<j} \) denotes the sequence of tokens preceding the \( j \)-th token in passage \( x \)  without conditioning on other sampled tokens.

We estimate the curvature at the point \( x \) by comparing the value of \( \hat{p}_\theta(x|x) \) with the values of perturbed texts \( \hat{p}_\theta(\tilde{x}|x) \). If \( \hat{p}_\theta(x|x) \) has a higher or equal value compared with the average of \( \hat{p}_\theta(\tilde{x}|x) \), the function has a positive curvature at \( x \), indicating that \( x \) is more likely human-written. Conversely, if \( \hat{p}_\theta(x|x) \) is lower, the curvature is negative, suggesting that \( x \) is more likely machine-revised.

Formally, following Fast-DetectGPT, we quantify the HLP-CPC as:
{\setlength{\belowdisplayskip}{2pt} 
\begin{equation}
    d(x, \hat{p}_\theta,  q_\phi ) = \frac{\log \hat{p}_\theta(x|x) - \mu}{\sigma},
\end{equation}
where
{\setlength{\belowdisplayskip}{2pt} 
\begin{align}
    \mu &= \mathbb{E}_{\tilde{x} \sim q_\phi(\tilde{x}|x)} \left[ \log \hat{p}_\theta(\tilde{x}|x) \right], \\
    \sigma^2 &= \mathbb{E}_{\tilde{x} \sim q_\phi(\tilde{x}|x)} \left[ \left( \log \hat{p}_\theta(\tilde{x}|x) - \mu \right)^2 \right],
\end{align}

We establish a detection criterion based on the HLP-CPC metric and a pre-defined threshold \( \epsilon \):
\begin{equation}
    f(x) =
    \begin{cases}
        1, & \text{if }  -d(x,  \hat{p}_\theta,  q_\phi )  > \epsilon, \\
        0, & \text{otherwise}.
    \end{cases}
\end{equation}
Here, \( f(x) = 1 \) indicates that the text is machine-revised, and \( f(x) = 0 \) indicates that the text is human-written. 
Note that we use the negative sign before \(d(x,\hat{p}_\theta,q_\phi) \) because our scoring model tends to assign higher scores to human-like text and lower scores to machine-revised text, making \( d(x) \) negative. By multiplying by \(-1\), we can more directly apply the threshold \(\epsilon\) to detect revised text.
\section{Experiments}
\label{sec:experiments}
\vspace{0pt}
To evaluate the efficacy and the potential application of our proposed HLPD method, we conduct experiments and comprehensive analysis of the results to answer the
following research questions:
\begin{enumerate}[leftmargin=*,label=\textbf{RQ\arabic*.},noitemsep, topsep=0pt, itemsep=0pt, partopsep=0pt]
    \item How effective and efficient of HLPD compare to other state-of-the-art (SOTA) methods?
    \item How robust of HLPD in detecting multi-task adversarial machine-revised text?
    \item How effective of our strategy HLPO compare to other baselines?
    \item How are the performance and generalizability of our designed Loss function?
\end{enumerate}
\subsection{Experiment Settings}
\label{subsec:datasets}
\noindent\textbf{Constructing the Single-Task Dataset.} 
Following ImBD's methodology for revision tasks and generation task, datasets were constructed based on four distinct tasks: Rewriting, Expand, Polish and Generate. The process involved a two-stage pipeline: (1) Revision instructions generation (2) Paragraph revision under generated instructions. See Appendix B.1, B.2 for details of all tasks.

\noindent\textbf{Constructing the Adversarial Multi-Task  Dataset.} To evaluate model robustness under realistic conditions, we constructed a challenging adversarial dataset inspired by Zhao et al. (2024). We developed a five-dimensional prompt generator that creates complex instructions by combining: (1) a core revision goal, (2) a target style, (3) instructions to add human-like qualities for detection evasion, (4) operational constraints, and (5) auxiliary user requests. Our generation pipeline consists of two stages: first, these five components are used to prompt DeepSeek-R1, which generates a consolidated instruction. Second, this instruction, along with a human-written text, is fed to a target LLM to produce the final adversarial sample. This methodology expanded our prompt pool to 750 unique prompts. The detailed pipeline is illustrated in Appendix B.3.

\noindent\textbf{Source Models.}
 To ensure a comprehensive evaluation that reflects realistic scenarios, we extended the benchmark's original models (GPT-3.5-Turbo, Qwen2-7B \citep{yang2024qwen}, Llama-3-8B \citep{touvron2023llama}, Mixtral-7B \citep{jiang2024mixtral}, Deepseek-7B \citep{deepseek2024b}, ) with more recent, advanced LLMs like GPT-o3, GPT-4o
 , Deepseek-R1 \cite{guo2025deepseek}, Gemini-2.5Pro \cite{GoogleDeepmind}, Grok-3 \cite{xai2024grok}, Claude-3.5 \cite{Anthropic}.

\noindent\textbf{Training datasets.}
To ensure a fair comparison, we constructed our training dataset using the same human-written source texts as the ImBD training set. Each text was then revised by GPT-3.5-Turbo, which randomly performed one of four revision tasks: expansion, polishing, rewriting, or generation, as explained above. The final dataset consists of pairs, each containing an original human-written text and its corresponding machine-revised version.


\noindent\textbf{Testing datasets.} 
We adopt the experimental setup from prior work \citep{bao2023fast, howard2018universal} and source our human-written text from five diverse datasets: \textit{XSum} \cite{narayan2018don} (news articles), \textit{SQuAD} \cite{rajpurkar2016squad} (question answering), \textit{WritingPrompts} (`Writing') \cite{fan2018hierarchical} (creative stories), \textit{PubMedQA} \cite{jin2019pubmedqa} (biomedical QA), and \textit{WikiText} (`Wiki')~\cite{merity2016pointer} (encyclopedic text).
\noindent\textbf{Baselines.}
We compare our method with two categories of approaches: five training-based models (ReMoDetect \cite{lee2024remodetect}, Ghostbuster \cite{verma-etal-2024-ghostbuster}, RoBERTa-base, RoBERTa-large \citep{liu2019roberta} and the commercial detector GPTZero \citep{tian2023gptzero} and nine logit-based models (ImBD 
\cite{tang2024codi}, Likelihood \citep{ippolito2019automatic}, LogRank \citep{solaiman2019release}, Entropy \citep{gehrmann2019gltr}, LRR \citep{su2023detectllm}, NPR \citep{su2023detectllm}, DNA-GPT \citep{yang2023dna}, DetectGPT and Fast-DetectGPT).

To ensure the robustness of our findings, all key experiments were repeated five times with different random seeds.Figure~\ref{pic:detection_roc_eps} presents the ROC curves for detection performance. The notation HLPD$\dag$ refers to a baseline version implemented without any optimizations. Further implementation and training details are provided in Appendix A. 

\vspace{0pt}
\begin{figure}[ht!]
    \centering
    \includegraphics[width=\linewidth]{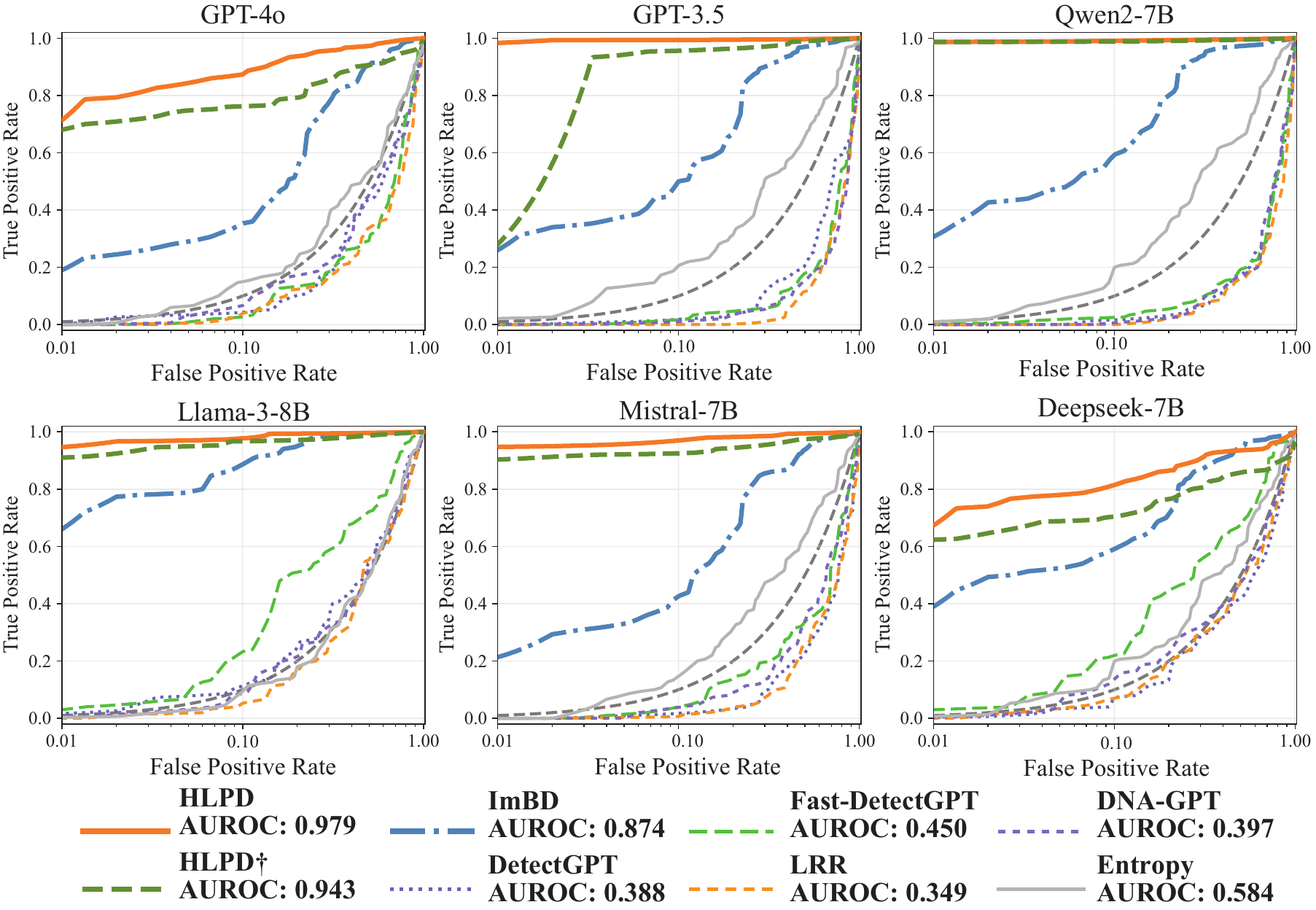}
    \caption{\textbf{ROC Curve for Detection.}}
    \label{pic:detection_roc_eps}
\end{figure}

\begin{table*}[h]
\setlength{\belowcaptionskip}{0pt}
\setlength{\floatsep}{0pt}     
\setlength{\textfloatsep}{0pt}  
\belowrulesep=0pt
\aboverulesep=0pt
\scriptsize
    \centering
    
    \resizebox{\textwidth}{!}{
    \begin{threeparttable}
    \setlength{\tabcolsep}{4pt} 
   \begin{tabular}{l|c|ccccc|ccccc|c}
    \toprule
    \multirow{2}{*}{\textbf{Method}} & \textbf{Time cost} & \multicolumn{5}{c|}{\textbf{GPT-3.5-turbo}} & \multicolumn{5}{c|}{\textbf{GPT-4o}}  & \multirow{2}{*}{ \textbf{\makecell[c]{Overall\\ Avg.} } } \\
    &  (s/1k words)& XSum & Writing & PubMed & Squad & Wiki  &XSum & Writing & PubMed & Squad & Wiki  &  \\
    \midrule
    \textbf{RoBERTa-base} & 0.07 & 0.5806 & 0.7225 & 0.4370 & 0.4588 & 0.5930  & 0.4921 & 0.4774 & 0.2496 & 0.4898 & 0.6512  & 0.5152 \\
    \textbf{RoBERTa-large} & 0.11 & 0.6391 & 0.7236 & 0.4848 & 0.4863 & 0.6028  & 0.4782 & 0.4708 & 0.3089 & 0.4716 & 0.6378 & 0.5304 \\
    \textbf{Entropy} & 0.35 & 0.6236 & 0.4564 & 0.5160 & 0.4942 & 0.3288  & 0.5351 & 0.3281 & 0.4923 & 0.4852 & 0.3074  & 0.4567 \\
    \textbf{Likelihood} & 0.38 & 0.2774 & 0.5448 & 0.4481 & 0.4814 & 0.6250 & 0.4290 & 0.6834 & 0.4955 & 0.5206 & 0.6950 & 0.5200 \\
    \textbf{Logrank} & 0.36 & 0.2528 & 0.4847 & 0.4454 & 0.4534 & 0.6097  & 0.4064 & 0.6581 & 0.4936 & 0.5008 & 0.6716  & 0.4976 \\
    \textbf{LRR} & 0.41 & 0.2185 & 0.3208 & 0.4505 & 0.3760 & 0.5312  & 0.3647 & 0.5528 & 0.4820 & 0.4422 & 0.5641  & 0.4303 \\
    \textbf{NPR} & 111.99 & 0.2873 & 0.5753 & 0.4118 & 0.4018 & 0.4822 & 0.4066 & 0.7067 & 0.4811 & 0.4980 & 0.5131  & 0.4764 \\
    \textbf{DNA-GPT} & 35.92 & 0.2381 & 0.5284 & 0.4139 & 0.4611 & 0.6252  & 0.4492 & 0.6204 & 0.4933 & 0.4759 & 0.6720  & 0.4978 \\
    \textbf{Detect-GPT} & 111.33 & 0.3118 & 0.6023 & 0.4320 & 0.4357 & 0.4629 & 0.4350 & 0.7270 & 0.4949 & 0.5195 & 0.5025  & 0.4924 \\ 
    \textbf{Fast-DetectGPT} & 0.72 & 0.2684 & 0.5520 & 0.4404 & 0.4399 & 0.4192  & 0.3963 & 0.6212 & 0.4842 & 0.4718 & 0.4944  & 0.4588 \\
    \textbf{ImBD} & 0.72 & 0.8651 & 0.8828 & 0.6218 & 0.9223 & 0.7930  & 0.7995 & 0.8136 & 0.6178 & 0.8396 & 0.7881 & 0.7870 \\
    \textbf{GPTZero} & 1.64 & 0.9204 & 0.8484 & 0.6453 & 0.8490 & 0.6794  & 0.8023 & 0.8081 &0.7640& 0.7548  & 0.7986 & 0.7944 \\
    \textbf{ReMoDetect} & 0.28 & 0.9013 & 0.8643 & 0.8343  & 0.8693 & 0.8293 & 0.7973 &\textbf{0.8553} &0.7547 &0.8133 &0.8493 &0.8368\\
\textbf{Ghostbuster}  &0.23&0.9350&	0.9728&0.8087	&0.9033	&0.9602	&	0.9295&	0.8306	&\textbf{0.7853}&0.8663&	0.8681	&	0.8860\\
      \rowcolor{gray!25}
    \textbf{HLPD } & 0.72 & \textbf{0.9998} & \textbf{0.9813} & \textbf{0.8125} & \textbf{0.9598} & \textbf{0.9796} &   \textbf{0.9619} & 0.8435 & 0.7393 &\textbf{0.9125} & \textbf{0.9549}  & \textbf{0.9144} \\
    \bottomrule
    \end{tabular}

    \end{threeparttable}
    }
\caption{\textbf{Detection Performance of GPT-3.5 and GPT-4o rewritten text.} Metric: AUROC. }
\label{tab:rewritten_text_detection}
  \vspace{0pt}  
\end{table*}

\vspace{0pt}
\subsection{Effectiveness \& Efficiency  (RQ1)}
\vspace{0pt}
\label{subsec:results}
\noindent\textbf{Detection Performance for Single-task Revision.} 
\label{subsec:results-rq2}
As shown in Table~\ref{tab:rewritten_text_detection}, in ImBD's rewrite task, our method outperforms ImBD by 15.11\% relative improvement in detecting GPT-series outputs. 
Compared to the ReMoDetect and Ghostbuster, our method achieves average improvements of 9.27\% and 3.21\%, respectively. 
Additionally, our method achieves a 10.57\% relative improvement over GPTZero. This indicates that our approach is highly training-efficient, attaining superior performance with a limited amount of data compared to models trained on significantly large datasets. See Appendix D Table~\ref{tab:polished_text_detection} 
for the performance on \textbf{polish} task. Furthermore, our method gets the highest score on four \textit{open-source} models.
See Appendix D Table~\ref{gptzeroc}, \ref{tab:detals on open source model polished}, and \ref{tab:detals on open source model rewritten} 
for detailed results. 

\begin{table}
  \centering
  \large 
  \vspace{-2pt}
  \setlength{\belowcaptionskip}{0pt}

  \renewcommand{\arraystretch}{1.1}
  \setlength{\tabcolsep}{4pt}

  \adjustbox{max width=\linewidth,keepaspectratio=false}{%
    \begin{tabular}{lccccc}
      \toprule
        \multirow{2}{*}{\textbf{Method}}
        & \multicolumn{4}{c}{\textbf{Source Model}}
        & \multirow{2}{*}{\textbf{Avg.}} \\
      \cmidrule(lr){2-5}
        & GPT‑o3 & DeepSeek‑R1 & Gemini‑2.5 & Grok‑3 & \\
      \midrule
      Fast‑DetectGPT & 0.8164 & 0.3765 & 0.6716 & 0.6649 & 0.6324 \\
      ImBD           & 0.9238 & 0.9114 & 0.8688 & 0.9873 & 0.9228 \\
      GPTZero        & 0.8266 & 0.6787 & 0.8061 & 0.6882 & 0.7499 \\
      \rowcolor{gray!20}
      \textbf{HLPD}  & \textbf{0.9780} & \textbf{0.9850} & \textbf{0.9453} & \textbf{0.9869} & \textbf{0.9738} \\
      \bottomrule
    \end{tabular}
  }
  
  \caption{\textbf{Performance of Advanced Methods Across Generation Tasks on Xsum Dataset.}}
  \label{tab:model_performance_tasks}
  \vspace{-5pt}
\end{table}

\noindent\textbf{Detecting Advanced LLM Generations.} As shown in Table~\ref{tab:model_performance_tasks},  HLPD demonstrates superior performance on text from advanced LLMs, outperforming ImBD by an average of 5.53\%. This suggests that HLPD's strategy of aligning with stable human language patterns, rather than shifting machine-generated ones, provides greater resilience to the architectural and stylistic variations inherent in newer models.

\noindent\textbf{Multilingual Detection Results.} To evaluate the performance of HLPD in multilingual text scenarios, we generated Chinese, Portuguese, and Spanish dataset variants. As shown in Appendix D Figure~\ref{pic:multilanguage}, HLPD exhibits superior detection capabilities: achieving a 10\% relative advantage over ImBD and gaining an average improvement of 33\% against RoBERTa-large across all language versions.

\noindent\textbf{Inference Efficiency.}
As shown in Table~\ref{tab:rewritten_text_detection}, HLPD achieves an average inference time of 0.72 seconds per 1,000 words, matching the efficiency of Fast-DetectGPT while delivering superior detection performance on L20 GPU cards.

\vspace{0pt}

\subsection{Robustness Studies(RQ2)}
\vspace{0pt}
\noindent\textbf{Robustness in  Adversarial Multi-Task Revisions.} As detailed in Table~\ref{tab:model_performance_tasks_left}, the efficacy of HLPD in detecting adversarial multi-task revised texts from four prominent LLMs was evaluated against leading baselines. HLPD achieved an overall average AUROC of 0.9463, surpassing ImBD by 13.52\% and outperforming ReMoDetect by 6.69\%. This indicated that HLPD's  robustness and enhanced ability to identify multi-task machine-revised text even when it employs sophisticated adversarial prompts.

\begin{table}[h]
\setlength{\belowcaptionskip}{0pt}
  \centering
  \small
  \vspace{-2pt}
  
  \renewcommand{\arraystretch}{1.2}
  \setlength{\tabcolsep}{4pt}
  \resizebox{\linewidth}{!}{
    \begin{tabular}{llccccc}
      \toprule
      \multirow{2}{*}{\textbf{Model}} 
      & \multirow{2}{*}{\textbf{Method}} 
      & \multicolumn{4}{c}{\textbf{Dataset}} 
      & \multirow{2}{*}{\textbf{Avg.}} \\
      \cmidrule(lr){3-6}
      & & Xsum & Writing & SQuAD & Wiki & \\
      \midrule
      \multirow{4}{*}{\textbf{GPT-4o}} 
      & Fast-DetectGPT & 0.3621 & 0.6280 & 0.4685 & 0.4896 & 0.4870 \\
      & ImBD & 0.8601 & 0.9144 & 0.7735 & 0.8285 & 0.8441 \\
      & GPTZero & 0.9133 & 0.7277 & 0.8630 & 0.7323 & 0.8091 \\
      & ReMoDetect & 0.8916 & 0.9016 & 0.8116 & 0.9016 & 0.8766 \\
      \rowcolor{gray!20}
      & \textbf{HLPD} & \textbf{0.9612} & \textbf{0.9564} & \textbf{0.9108} & \textbf{0.9572} & \textbf{0.9464} \\
      \midrule
      \multirow{4}{*}{\textbf{DeepSeek-R1}} 
      & Fast-DetectGPT & 0.2061 & 0.3612 & 0.3168 & 0.2924 & 0.2941 \\
      & ImBD & 0.8745 & 0.8476 & 0.7520 & 0.7513 & 0.8064 \\
      & GPTZero & 0.7997 & 0.6741 & 0.8336 & 0.7407 & 0.7620 \\
      & ReMoDetect & 0.9201 & 0.9041 & 0.9281 & 0.8961 & 0.9121 \\
      \rowcolor{gray!20}
      & \textbf{HLPD} & \textbf{0.9730} & \textbf{0.9281} & \textbf{0.9287} & \textbf{0.9708} & \textbf{0.9501} \\
      \midrule
      \multirow{4}{*}{\textbf{Claude-3.5}} 
      & Fast-DetectGPT & 0.2474 & 0.4572 & 0.4103 & 0.3722 & 0.3718 \\
      & ImBD & 0.9105 & 0.9098 & 0.8110 & 0.8504 & 0.8704 \\
      & GPTZero & 0.8266 & 0.6787 & 0.8061 & 0.6882 & 0.7499 \\
      & ReMoDetect & 0.9086 & 0.8866 &  0.9196 & 0.8756 & 0.8976 \\
      \rowcolor{gray!20}
      & \textbf{HLPD} & \textbf{0.9988} & \textbf{0.9535} & \textbf{0.9383} & \textbf{0.9512} & \textbf{0.9604} \\
      \midrule
      \multirow{4}{*}{\textbf{Gemini-2.5-Pro}} 
      & Fast-DetectGPT & 0.3148 & 0.5413 & 0.4525 & 0.3412 & 0.4125 \\
      & ImBD & 0.8544 & 0.9120 & 0.7599 & 0.7284 & 0.8137 \\
      & GPTZero & 0.8020 & 0.7060 & 0.7915 & 0.5745 & 0.7185 \\
      & ReMoDetect & 0.9416 & 0.7816 & \textbf{0.9016} & 0.8216  & 0.8616\\
      \rowcolor{gray!20}
      & \textbf{HLPD} & \textbf{0.9433} & \textbf{0.9328} & 0.8980 & \textbf{0.9394} & \textbf{0.9284} \\
      \bottomrule
    \end{tabular}
  }
  \caption{\textbf{Performance on Adversarial Multi-Task Revision.}}
  \label{tab:model_performance_tasks_left}
  \vspace{-5pt}
\end{table}


\noindent\textbf{Robustness in Diverse Revision Goals.} 
HLPD's robustness against diverse revision goals is evident in Appendix D Table~\ref{tab:diverse_tasks}. Evaluated on the XSum dataset revised by six LLMs, HLPD achieved the highest average AUROC across all three revision tasks. This performance represents a significant 32.38\% improvement over Fast-DetectGPT and a 4.11\% relative gain over ImBD, demonstrating HLPD's strong capability in multi-task revision and multi-context detection. For further details, see Appendix D Table~\ref{tab:performance across different Models and Tasks.}.

\noindent\textbf{Robustness in Multiple Revision Iterations.}
To assess HLPD's robustness to multiple revision iterations, we evaluated its performance on texts undergoing one to five revisions from two LLMs: GPT-4o and Gemini-2.5-Pro. As shown in Appendix D Table~\ref{tab:multi_iteration_enhanced}, HLPD consistently maintained highest average AUROC scores, indicating HLPD's detection efficacy even as texts undergo extensive iterative machine revisions.

\subsection{Ablation Studies (RQ3 \& RQ4)}
\label{Ablation}
\vspace{0pt}
\noindent\textbf{Ablation on Strategy.}
As detailed in Table~\ref{tab:detection_strategies}, our method, HLPD, achieves a remarkable average AUROC improvement of 56\% when benchmarked against Fast-DetectGPT, a baseline established without alignment techniques. HLPD further demonstrates consistent superiority by outperforming SFT, RLHF,  ORPO and IPO by average margins of 30\%, 26\%, 23\% and 23\%, respectively. Notably, our optimized HLPO$\ddagger\star$ variant secures a 14.5\% relative advantage over ImBD, strongly affirming the efficacy of detection paradigms aligned with human writing styles.

\noindent\textbf{Ablation on Optimization.}
To demonstrate the efficacy of our innovations, as explained in Section \ref{subsec:human_style_preference_optimization}, we ablated the core components of our loss function: the switch to a linear loss (from sigmoid activation) and the introduction of a dynamically adjusted $\beta$ (compared to a fixed beta from DPO/ImBD). As detailed in \cref{tab:detection_strategies}, isolating the dynamic $\beta$ (HLPO$\star$) resulted in a 1\% relative accuracy improvement. The linear loss function alone (HLPO$\ddagger$) contributed a more substantial 5\% relative improvement. The integration of these two strategies in HLPO ($\ddagger\star$)  yielded a relative improvement of 8\%, indicating a  synergistic effect.This benefit extended to the ImBD baseline, where ImBD$\ddagger\star$ outperformed ImBD by 6\%. These results confirm that our modifications are robustly beneficial, offering enhanced training performance, particularly when data or training epochs are limited.

\begin{table}[htbp!]
    \centering
    \scriptsize
    
    \resizebox{\linewidth}{!}{
    \begin{threeparttable}
        \setlength{\tabcolsep}{2pt}
        \begin{tabular}{lccccccccc}
            \toprule
            \multirow{2}{*}{\textbf{Strategy}} & \multicolumn{4}{c}{\textbf{Deepseek-R1}} 
            &\multicolumn{4}{c}{\textbf{Gemini-2.5-Pro}} &\multirow{2}{*}{\textbf{Avg.}} \\
            \cmidrule(lr){2-5} \cmidrule(lr){6-9}
            & XSum & Writ.  & SQu. & Wiki  & XSum & Writ.  & SQu. & Wiki& \\
            \midrule
            \textbf{W/O A.} & 0.21 & 0.36 & 0.32 & 0.29 &0.31&0.54&0.45&0.34& 0.35  \\
            \textbf{SFT} & 0.68 & 0.70 & 0.66 & 0.78 &0.67&0.73&0.65&0.81& 0.71 \\
            \textbf{ORPO} & 0.75 & 0.70 & 0.67 & 0.79 &0.77&0.74&0.69&0.80&0.74 \\
            \textbf{IPO} & 0.75 & 0.70 & 0.67 & 0.79 &  0.71 &0.78&0.61&0.67&0.71\\
            \textbf{RLHF} & 0.70 & 0.82 & 0.78 & 0.54 & 0.64&0.79&0.72&0.68& 0.71\\
            \hline
            \textbf{ImBD} & 0.87 & 0.84 & 0.75 & 0.75&0.85&0.91&0.76& 0.73& 0.81 \\
            \textbf{ImBD$\ddagger\star$} & 0.81 & 0.92 & 0.79 & 0.85 &0.87&0.92&0.87&0.85& 0.86 \\
            \textbf{HLPO}  & 0.93 & 0.85 & 0.89 & 0.87 &0.95&0.87&0.81&0.89& 0.88 \\
            \textbf{HLPO$\star$}  & 0.93 & 0.89 & 0.90 & 0.91 & 0.92&0.86 & 0.85&0.86 &0.89 \\
            \textbf{HLPO$\ddagger$}  & 0.92 & 0.90 & 0.89 & 0.86 &\textbf{0.95}&0.87&\textbf{0.93}&0.94& 0.91 \\
            \textbf{HLPO$\ddagger\star$}  & \textbf{0.97} & \textbf{0.93} & \textbf{0.93} & \textbf{0.97} & 0.94&\textbf{0.93}&0.90&\textbf{0.94}&\textbf{0.94} \\
            \bottomrule
        \end{tabular}
        \begin{tablenotes}
            \scriptsize
            \item[1] SFT
            \citep{ziegler2019fine}, RLHF\citep{christiano2017deep}, and ORPO\citep{hong2024orpo}, IPO\cite{pmlr-v238-gheshlaghi-azar24a} are different preference strategies. ``W/O A.''denotes training with non-alignment or preference optimization.
            \item[2] $\ddagger$ denotes the loss function incorporates our proposed linearization of HLPO; $\star$ denotes the loss function is scaled by a dynamically adjusted  $\beta$ to improve robustness.
        \end{tablenotes}
    \end{threeparttable}
    }
    \caption{\textbf{Ablation on Strategy \& Optimization for Adversarial Multi-Task Revisions.}}
    \label{tab:detection_strategies}
    \vspace{-8pt}
\end{table}
\vspace{-3pt}
\subsection{Humanization Adaptive Attack Experiments}
\label{subsec:humanization_experiments}
\vspace{0pt}

\label{subsec:humanizing_machine_texts}
\vspace{0pt}
After HLPO, we get a model that prefers the human language style. This scoring model shows its potential in a fair amount of downstream tasks besides detection. Here, we present a case that uses an additional strategy to make machine-revised texts less detectable.
We apply the perturbation model making minor edits to introduce variability of human writing, and the scoring model after HLPO to pick up the most human-looked edited text. The process involves:
\textbf{1) Perturbation Generation}: For each machine-revised text \( x_m \), we generate perturbed versions \( \tilde{x}_m \) by sampling from \( q_\phi(\tilde{x}|x_m) \), introducing human-style variations.
\textbf{2) Selection of Humanized Texts}: Among the perturbed texts, we select those with higher log-probabilities under the scoring model \( \hat{p}_\theta \), as they are more likely to align with human writing styles.
\textbf{3) Replacement}: Replace the original machine-revised text \( x_m \) with the selected \( \tilde{x}_m \) to obtain a humanized version.

The selected perturbation \( \tilde{x}_m \) can be applied to additional iterations for further reductions in detection probability. However, it is important to note that repeating this process might introduce more substantial semantic deviations from the original text. Please see Appendix C for the process flow diagram.

\noindent\textbf{Experiment Settings.} 
We generate our text samples via a two-stage process. First, to create the base machine-revised text, we use GPT-3.5-Turbo to polish 100 paragraphs sourced from the XSum dataset. Second, for each revised paragraph, we employ a perturbation-and-selection method: T3-B5 generates 100 candidate perturbations, and our scoring model selects the one with the highest preference score as the final ``humanized'' output for each humanization. 

\noindent\textbf{Effectiveness Against Commercial Detectors.}
As shown in Appendix C Figure~\ref{fig:humanize-perf}, our method reduces the detection AUROC by an average of 5\% per iteration, culminating in a 20\% total reduction after four iterations compared to the original machine-revised text. On the commercial GPTZero platform, this corresponds to a significant 74\% decrease in the predicted AI probability. These results highlight the efficacy of our iterative humanization technique in evading commercial detection systems.

\noindent\textbf{Effectiveness Against Open-Source Detectors.}
The method demonstrates similar effectiveness against open-source detectors. On the ImBD demo, our 4-iteration humanized text reduced the AI probability score by 44.5\% (from 81.8\% to 37.3\%). Similarly, on the Ghostbuster demo, the probability dropped by 61\% (from 87\% to 26\%). See detailed examples of the original and humanized texts in Appendix C.3.

\vspace{-5pt}
\section{Related Work}
We classify current detection methodologies into three primary paradigms, each with distinct limitations in discerning human-machine hybrid content.

\noindent\textbf{Non-Alignment Detection.} These approaches use either classifier training or logit-based analysis. Supervised models \cite{liu2019roberta} achieve domain-specific accuracy but suffer from distributional overfitting. Probability-based metrics \cite{mitchell2023detectgpt,liu2019roberta}  analyze token-level characteristics through log-likelihood and curvature estimation,
While methods like DetectGPT \cite{mitchell2023detectgpt} enable zero-shot detection, they remain insensitive to style-preserving machine edits.

\noindent\textbf{Machine-Aligned Detection.} These new methods like ImBD employ style preference optimization on the scoring model to capture machine-generated patterns. Though effective for overt AI features, their imitation-based paradigm demonstrates limited discriminative power when confronting more advanced LLMs under diverse, adversarial prompts. 

\noindent\textbf{Human-Aligned Models.}
In NLP, the goal of aligning models with human text is a major focus, with efforts centered on making generated texts more human-like or better matching human intent. InstructGPT \cite{ouyang2022training} and RLHF \cite{christiano2017deep} use human feedback to guide models toward producing more natural, human-understandable texts.
Additionally, Chain-of-Thought (CoT) \cite{wei2022chain} prompting improves model reasoning by guiding the model to break down tasks into logical steps, allowing it to generate more interpretable and human-like output. However, these techniques still focus mainly on improving generation rather than capturing the difference between human-style and machine-style texts.
Our HLPD (Human-Language-Preference-Detection) framework enlarges this gap by enabling human language style alignment. Unlike ImBD, HLPD trains the model towards a completely opposite direction for scoring, 
shows a 13.52\% improvement in multi-task adversarial revisions  and achieves a 5.53\% higher AUROC in machine-generated text across four advanced LLMs compared to ImBD. Also, the trained scoring model shows some extra potential in downstream tasks like adaptive attack.
\section{Conclusion and Limitations}
\label{sec:conclusion}
\vspace{0pt}

We presented HLPD, which aligns scoring models with human language style, achieving significant accuracy improvements in detecting both fully machine-generated and multi-task revised texts under black-box conditions. HLPD's robustness and adaptability were validated through our novel adversarial multi-task evaluation framework, featuring a five-dimensional prompt generator and diverse LLMs. This framework confirmed HLPD's superiority over existing methods across various tasks and languages. The importance of preference optimization was underscored by ablation studies. This work advances text detection through style alignment, paving the way for more resilient systems, and demonstrated HLPD's wider potential in an adaptive attack on GPTZero.
As for limitations, despite extensive evaluation across multiple state-of-the-art LLMs and diverse datasets, it remains unclear how the proposed detection approach will generalize to other models and domains not covered in our experiments. Additionally, our method may exhibit degraded performance when applied to very short sentences, where limited contextual information constrains reliable style and coherence based detection. 

\section*{Acknowledgments}
We thank all the anonymous reviewers for their constructive feedback. The authors are supported by NSFC (62502281), Shandong Provincial Natural Science Foundation (ZR2025QC1560), Jiangsu Provincial Natural Science Foundation (BK20250411), and Taishan Scholars Program.

\bibliography{references}

\begin{thebibliography}{50}
\providecommand{\natexlab}[1]{#1}

\bibitem[{Achiam et~al.(2023)Achiam, Adler, Agarwal, Ahmad, Akkaya, Aleman, Almeida, Altenschmidt, Altman, Anadkat et~al.}]{achiam2023gpt}
Achiam, J.; Adler, S.; Agarwal, S.; Ahmad, L.; Akkaya, I.; Aleman, F.~L.; Almeida, D.; Altenschmidt, J.; Altman, S.; Anadkat, S.; et~al. 2023.
\newblock Gpt-4 technical report.
\newblock \emph{arXiv preprint arXiv:2303.08774}.

\bibitem[{AlAfnan and MohdZuki(2023)}]{alafnan2023artificial}
AlAfnan, M.~A.; and MohdZuki, S.~F. 2023.
\newblock Do artificial intelligence chatbots have a writing style? An investigation into the stylistic features of ChatGPT-4.
\newblock \emph{Journal of Artificial intelligence and technology}, 3(3): 85--94.

\bibitem[{Anthropic(2025)}]{Anthropic}
Anthropic. 2025.
\newblock Cluade.
\newblock \url{https://claude.ai}.

\bibitem[{Bao et~al.(2023)Bao, Zhao, Teng, Yang, and Zhang}]{bao2023fast}
Bao, G.; Zhao, Y.; Teng, Z.; Yang, L.; and Zhang, Y. 2023.
\newblock Fast-detectgpt: Efficient zero-shot detection of machine-generated text via conditional probability curvature.
\newblock \emph{arXiv preprint arXiv:2310.05130}.

\bibitem[{Bian et~al.(2024)Bian, Lin, Liu, Lu, Zhang, He, Han, and Sun}]{bian2024influence}
Bian, N.; Lin, H.; Liu, P.; Lu, Y.; Zhang, C.; He, B.; Han, X.; and Sun, L. 2024.
\newblock Influence of external information on large language models mirrors social cognitive patterns.
\newblock \emph{IEEE Transactions on Computational Social Systems}.

\bibitem[{Bradley and Terry(1952)}]{bradley1952rank}
Bradley, R.~A.; and Terry, M.~E. 1952.
\newblock Rank analysis of incomplete block designs: I. The method of paired comparisons.
\newblock \emph{Biometrika}, 39(3/4): 324--345.

\bibitem[{Chen et~al.(2024)Chen, Zhu, Liu, Chen, Chen, Yuan, Leong, Li, Long, Zhang, Yan, Mei, Zhang, and Zhang}]{tang2024codi}
Chen, J.; Zhu, X.; Liu, T.; Chen, Y.; Chen, X.; Yuan, Y.; Leong, C.~T.; Li, Z.; Long, T.; Zhang, L.; Yan, C.; Mei, G.; Zhang, J.; and Zhang, L. 2024.
\newblock Imitate Before Detect: Aligning Machine Stylistic Preference for Machine-Revised Text Detection.

\bibitem[{Christiano et~al.(2017)Christiano, Leike, Brown, Martic, Legg, and Amodei}]{christiano2017deep}
Christiano, P.~F.; Leike, J.; Brown, T.; Martic, M.; Legg, S.; and Amodei, D. 2017.
\newblock Deep reinforcement learning from human preferences.
\newblock \emph{Advances in neural information processing systems}, 30.

\bibitem[{de~Lima-Santos and Ceron(2021)}]{de2021artificial}
de~Lima-Santos, M.-F.; and Ceron, W. 2021.
\newblock Artificial intelligence in news media: current perceptions and future outlook.
\newblock \emph{Journalism and media}, 3(1): 13--26.

\bibitem[{Deepmind(2024)}]{GoogleDeepmind}
Deepmind, G. 2024.
\newblock Gemini - Google DeepMind.
\newblock \url{https://deepmind.google/technologies/gemini/}.

\bibitem[{{DeepSeek-AI} et~al.(2024)}]{deepseek2024b}
{DeepSeek-AI}; et~al. 2024.
\newblock Deepseek-7B: A High-Performance Autoregressive Language Model for Natural Language Processing.
\newblock \emph{arXiv preprint arXiv:2403.01987}.

\bibitem[{Fan, Lewis, and Dauphin(2018)}]{fan2018hierarchical}
Fan, A.; Lewis, M.; and Dauphin, Y.~N. 2018.
\newblock Hierarchical Neural Story Generation.
\newblock In \emph{Proceedings of the 56th Annual Meeting of the Association for Computational Linguistics}, 889--898.

\bibitem[{Fang et~al.(2023)Fang, Ng, Leung, and Chu}]{fang2023systematic}
Fang, X.; Ng, D. T.~K.; Leung, J. K.~L.; and Chu, S. K.~W. 2023.
\newblock A systematic review of artificial intelligence technologies used for story writing.
\newblock \emph{Education and Information Technologies}, 28(11): 14361--14397.

\bibitem[{Ferrara(2023)}]{ferrara2023should}
Ferrara, E. 2023.
\newblock Should chatgpt be biased? challenges and risks of bias in large language models.
\newblock \emph{arXiv preprint arXiv:2304.03738}.

\bibitem[{Gehrmann, Strobelt, and Rush(2019)}]{gehrmann2019gltr}
Gehrmann, S.; Strobelt, H.; and Rush, A.~M. 2019.
\newblock Gltr: Statistical detection and visualization of generated text.
\newblock \emph{arXiv preprint arXiv:1906.04043}.

\bibitem[{Gheshlaghi~Azar et~al.(2024)Gheshlaghi~Azar, Daniel~Guo, Piot, Munos, Rowland, Valko, and Calandriello}]{pmlr-v238-gheshlaghi-azar24a}
Gheshlaghi~Azar, M.; Daniel~Guo, Z.; Piot, B.; Munos, R.; Rowland, M.; Valko, M.; and Calandriello, D. 2024.
\newblock A General Theoretical Paradigm to Understand Learning from Human Preferences.
\newblock In Dasgupta, S.; Mandt, S.; and Li, Y., eds., \emph{Proceedings of The 27th International Conference on Artificial Intelligence and Statistics}, volume 238 of \emph{Proceedings of Machine Learning Research}, 4447--4455. PMLR.

\bibitem[{Guo et~al.(2023)Guo, Zhang, Wang, Jiang, Nie, Ding, Yue, and Wu}]{guo2023close}
Guo, B.; Zhang, X.; Wang, Z.; Jiang, M.; Nie, J.; Ding, Y.; Yue, J.; and Wu, Y. 2023.
\newblock How close is chatgpt to human experts? comparison corpus, evaluation, and detection.
\newblock \emph{arXiv preprint arXiv:2301.07597}.

\bibitem[{Guo et~al.(2025)Guo, Yang, Zhang, Song, Zhang, Xu, Zhu, Ma, Wang, Bi et~al.}]{guo2025deepseek}
Guo, D.; Yang, D.; Zhang, H.; Song, J.; Zhang, R.; Xu, R.; Zhu, Q.; Ma, S.; Wang, P.; Bi, X.; et~al. 2025.
\newblock Deepseek-r1: Incentivizing reasoning capability in llms via reinforcement learning.
\newblock \emph{arXiv preprint arXiv:2501.12948}.

\bibitem[{Herbold et~al.(2023)Herbold, Hautli-Janisz, Heuer, Kikteva, and Trautsch}]{herbold2023large}
Herbold, S.; Hautli-Janisz, A.; Heuer, U.; Kikteva, Z.; and Trautsch, A. 2023.
\newblock A large-scale comparison of human-written versus ChatGPT-generated essays.
\newblock \emph{Scientific reports}, 13(1): 18617.

\bibitem[{Hong, Lee, and Thorne(2024)}]{hong2024orpo}
Hong, J.; Lee, N.; and Thorne, J. 2024.
\newblock Orpo: Monolithic preference optimization without reference model.
\newblock In \emph{Proceedings of the 2024 Conference on Empirical Methods in Natural Language Processing}, 11170--11189.

\bibitem[{Howard and Ruder(2018)}]{howard2018universal}
Howard, J.; and Ruder, S. 2018.
\newblock Universal language model fine-tuning for text classification.
\newblock \emph{arXiv preprint arXiv:1801.06146}.

\bibitem[{Ippolito et~al.(2019)Ippolito, Duckworth, Callison-Burch, and Eck}]{ippolito2019automatic}
Ippolito, D.; Duckworth, D.; Callison-Burch, C.; and Eck, D. 2019.
\newblock Automatic detection of generated text is easiest when humans are fooled.
\newblock \emph{arXiv preprint arXiv:1911.00650}.

\bibitem[{Jiang, Wang et~al.(2024)}]{jiang2024mixtral}
Jiang, H.; Wang, Y.; et~al. 2024.
\newblock Mixtral: Combining Multiple Training Paradigms for Enhanced Autoregressive Models.
\newblock \emph{arXiv preprint arXiv:2402.11234}.

\bibitem[{Jin et~al.(2019)Jin, Dhingra, Liu, Cohen, and Lu}]{jin2019pubmedqa}
Jin, Q.; Dhingra, B.; Liu, Z.; Cohen, W.~W.; and Lu, X. 2019.
\newblock PubMedQA: A Dataset for Biomedical Research Question Answering.
\newblock \emph{arXiv preprint arXiv:1909.06146}.

\bibitem[{Kelley et~al.(2021)Kelley, Yang, Heldreth, Moessner, Sedley, Kramm, Newman, and Woodruff}]{kelley2021exciting}
Kelley, P.~G.; Yang, Y.; Heldreth, C.; Moessner, C.; Sedley, A.; Kramm, A.; Newman, D.~T.; and Woodruff, A. 2021.
\newblock Exciting, useful, worrying, futuristic: Public perception of artificial intelligence in 8 countries.
\newblock In \emph{Proceedings of the 2021 AAAI/ACM Conference on AI, Ethics, and Society}, 627--637.

\bibitem[{Lee, Tack, and Shin(2024)}]{lee2024remodetect}
Lee, H.; Tack, J.; and Shin, J. 2024.
\newblock ReMoDetect: Reward Models Recognize Aligned LLM's Generations.
\newblock \emph{arXiv preprint arXiv:2405.17382}.

\bibitem[{Liu(2019)}]{liu2019roberta}
Liu, Y. 2019.
\newblock Roberta: A robustly optimized bert pretraining approach.
\newblock \emph{arXiv preprint arXiv:1907.11692}.

\bibitem[{M~Alshater(2022)}]{m2022exploring}
M~Alshater, M. 2022.
\newblock Exploring the role of artificial intelligence in enhancing academic performance: A case study of ChatGPT.
\newblock \emph{Available at SSRN 4312358}.

\bibitem[{McKenna et~al.(2023)McKenna, Li, Cheng, Hosseini, Johnson, and Steedman}]{mckenna2023sources}
McKenna, N.; Li, T.; Cheng, L.; Hosseini, M.~J.; Johnson, M.; and Steedman, M. 2023.
\newblock Sources of hallucination by large language models on inference tasks.
\newblock \emph{arXiv preprint arXiv:2305.14552}.

\bibitem[{Merity et~al.(2016)Merity, Xiong, Bradbury, and Socher}]{merity2016pointer}
Merity, S.; Xiong, C.; Bradbury, J.; and Socher, R. 2016.
\newblock Pointer Sentinel Mixture Models.
\newblock \emph{arXiv preprint arXiv:1609.07843}.

\bibitem[{Mitchell et~al.(2023)Mitchell, Lee, Khazatsky, Manning, and Finn}]{mitchell2023detectgpt}
Mitchell, E.; Lee, Y.; Khazatsky, A.; Manning, C.~D.; and Finn, C. 2023.
\newblock Detectgpt: Zero-shot machine-generated text detection using probability curvature.
\newblock In \emph{International Conference on Machine Learning}, 24950--24962. PMLR.

\bibitem[{Narayan, Cohen, and Lapata(2018)}]{narayan2018don}
Narayan, S.; Cohen, S.~B.; and Lapata, M. 2018.
\newblock Don't Give Me the Details, Just the Summary! Topic-Aware Convolutional Neural Networks for Extreme Summarization.
\newblock In \emph{Proceedings of the 2018 Conference on Empirical Methods in Natural Language Processing}, 1797--1807.

\bibitem[{OpenAI(2022)}]{openai2022chatgpt}
OpenAI, T. 2022.
\newblock Chatgpt: Optimizing language models for dialogue. OpenAI.

\bibitem[{Ouyang et~al.(2022)Ouyang, Wu, Jiang, Almeida, Wainwright, Mishkin, Zhang, Agarwal, Slama, Ray et~al.}]{ouyang2022training}
Ouyang, L.; Wu, J.; Jiang, X.; Almeida, D.; Wainwright, C.; Mishkin, P.; Zhang, C.; Agarwal, S.; Slama, K.; Ray, A.; et~al. 2022.
\newblock Training language models to follow instructions with human feedback.
\newblock \emph{Advances in neural information processing systems}, 35: 27730--27744.

\bibitem[{Pu and Demberg(2023)}]{pu2023chatgpt}
Pu, D.; and Demberg, V. 2023.
\newblock ChatGPT vs human-authored text: Insights into controllable text summarization and sentence style transfer.
\newblock \emph{arXiv preprint arXiv:2306.07799}.

\bibitem[{Rafailov et~al.(2024)Rafailov, Sharma, Mitchell, Manning, Ermon, and Finn}]{rafailov2024direct}
Rafailov, R.; Sharma, A.; Mitchell, E.; Manning, C.~D.; Ermon, S.; and Finn, C. 2024.
\newblock Direct preference optimization: Your language model is secretly a reward model.
\newblock \emph{Advances in Neural Information Processing Systems}, 36.

\bibitem[{Rajpurkar et~al.(2016)Rajpurkar, Zhang, Lopyrev, and Liang}]{rajpurkar2016squad}
Rajpurkar, P.; Zhang, J.; Lopyrev, K.; and Liang, P. 2016.
\newblock SQuAD: 100,000+ Questions for Machine Comprehension of Text.
\newblock \emph{arXiv preprint arXiv:1606.05250}.

\bibitem[{Reinhart et~al.(2025)Reinhart, Markey, Laudenbach, Pantusen, Yurko, Weinberg, and Brown}]{reinhart2025llms}
Reinhart, A.; Markey, B.; Laudenbach, M.; Pantusen, K.; Yurko, R.; Weinberg, G.; and Brown, D.~W. 2025.
\newblock Do LLMs write like humans? Variation in grammatical and rhetorical styles.
\newblock \emph{Proceedings of the National Academy of Sciences}, 122(8): e2422455122.

\bibitem[{Solaiman et~al.(2019)Solaiman, Brundage, Clark, Askell, Herbert-Voss, Wu, Radford, Krueger, Kim, Kreps et~al.}]{solaiman2019release}
Solaiman, I.; Brundage, M.; Clark, J.; Askell, A.; Herbert-Voss, A.; Wu, J.; Radford, A.; Krueger, G.; Kim, J.~W.; Kreps, S.; et~al. 2019.
\newblock Release strategies and the social impacts of language models.
\newblock \emph{arXiv preprint arXiv:1908.09203}.

\bibitem[{Su et~al.(2023)Su, Zhuo, Wang, and Nakov}]{su2023detectllm}
Su, J.; Zhuo, T.~Y.; Wang, D.; and Nakov, P. 2023.
\newblock Detectllm: Leveraging log rank information for zero-shot detection of machine-generated text.
\newblock \emph{arXiv preprint arXiv:2306.05540}.

\bibitem[{Tian and Cui(2023)}]{tian2023gptzero}
Tian, E.; and Cui, A. 2023.
\newblock GPTZero: Towards detection of AI-generated text using zero-shot and supervised methods.
\newblock \emph{GPTZero}.

\bibitem[{Touvron et~al.(2023)Touvron, Martin, Stone, Albert, Almahairi, Babaei, Bashlykov, Batra, Bhargava, Bhosale et~al.}]{touvron2023llama}
Touvron, H.; Martin, L.; Stone, K.; Albert, P.; Almahairi, A.; Babaei, Y.; Bashlykov, N.; Batra, S.; Bhargava, P.; Bhosale, S.; et~al. 2023.
\newblock Llama 2: Open foundation and fine-tuned chat models.
\newblock \emph{arXiv preprint arXiv:2307.09288}.

\bibitem[{Tufts, Zhao, and Li(2024)}]{tufts2024practical}
Tufts, B.; Zhao, X.; and Li, L. 2024.
\newblock A practical examination of AI-generated text detectors for large language models.
\newblock \emph{arXiv preprint arXiv:2412.05139}.

\bibitem[{Verma et~al.(2024)Verma, Fleisig, Tomlin, and Klein}]{verma-etal-2024-ghostbuster}
Verma, V.; Fleisig, E.; Tomlin, N.; and Klein, D. 2024.
\newblock Ghostbuster: Detecting Text Ghostwritten by Large Language Models.
\newblock In Duh, K.; Gomez, H.; and Bethard, S., eds., \emph{Proceedings of the 2024 Conference of the North American Chapter of the Association for Computational Linguistics: Human Language Technologies (Volume 1: Long Papers)}, 1702--1717. Mexico City, Mexico: Association for Computational Linguistics.

\bibitem[{Wei et~al.(2022)Wei, Wang, Schuurmans, Bosma, Xia, Chi, Le, Zhou et~al.}]{wei2022chain}
Wei, J.; Wang, X.; Schuurmans, D.; Bosma, M.; Xia, F.; Chi, E.; Le, Q.~V.; Zhou, D.; et~al. 2022.
\newblock Chain-of-thought prompting elicits reasoning in large language models.
\newblock \emph{Advances in neural information processing systems}, 35: 24824--24837.

\bibitem[{{x.ai}(2024)}]{xai2024grok}
{x.ai}. 2024.
\newblock Grok.
\newblock \url{https://x.ai/grok}.
\newblock Accessed: 2024-12-29.

\bibitem[{Yang, Chen et~al.(2024)}]{yang2024qwen}
Yang, L.; Chen, F.; et~al. 2024.
\newblock Qwen-2: Advanced Autoregressive Models with Efficient Training Techniques.
\newblock \emph{arXiv preprint arXiv:2401.09876}.

\bibitem[{Yang et~al.(2023)Yang, Cheng, Wu, Petzold, Wang, and Chen}]{yang2023dna}
Yang, X.; Cheng, W.; Wu, Y.; Petzold, L.; Wang, W.~Y.; and Chen, H. 2023.
\newblock Dna-gpt: Divergent n-gram analysis for training-free detection of gpt-generated text.
\newblock \emph{arXiv preprint arXiv:2305.17359}.

\bibitem[{Yuan et~al.(2024)Yuan, Yuan, Tan, Wang, Huang, and Huang}]{yuan2024rrhf}
Yuan, H.; Yuan, Z.; Tan, C.; Wang, W.; Huang, S.; and Huang, F. 2024.
\newblock RRHF: Rank responses to align language models with human feedback.
\newblock \emph{Advances in Neural Information Processing Systems}, 36.

\bibitem[{Ziegler et~al.(2019)Ziegler, Stiennon, Wu, Brown, Radford, Amodei, Christiano, and Irving}]{ziegler2019fine}
Ziegler, D.~M.; Stiennon, N.; Wu, J.; Brown, T.~B.; Radford, A.; Amodei, D.; Christiano, P.; and Irving, G. 2019.
\newblock Fine-tuning language models from human preferences.
\newblock \emph{arXiv preprint arXiv:1909.08593}.

\end{thebibliography}

\clearpage
\newpage

\appendix

\onecolumn
\renewcommand{\appendixname}{Appendix~\Alph{Humanization performance on ImBD demo website}}


\section{Experimental Details}
In this section, we describe the experimental details of Experiments Section , including HLPD and baselines. To assess the stability of our results, we report 95\% confidence intervals (CIs) for our main multi-task adversarial experiments. For the results on the four datasets revised by Gemini-2.5 Pro, we conducted five independent runs using different random seeds (42, 199, 410, 2231, and 2533). The reported results present the mean Area Under the Receiver Operating Characteristic curve (AUROC) across these runs, with the error bars representing the 95\% CI.

\begin{figure}[h]
    \centering
    \includegraphics[width=0.48\linewidth,valign=t]{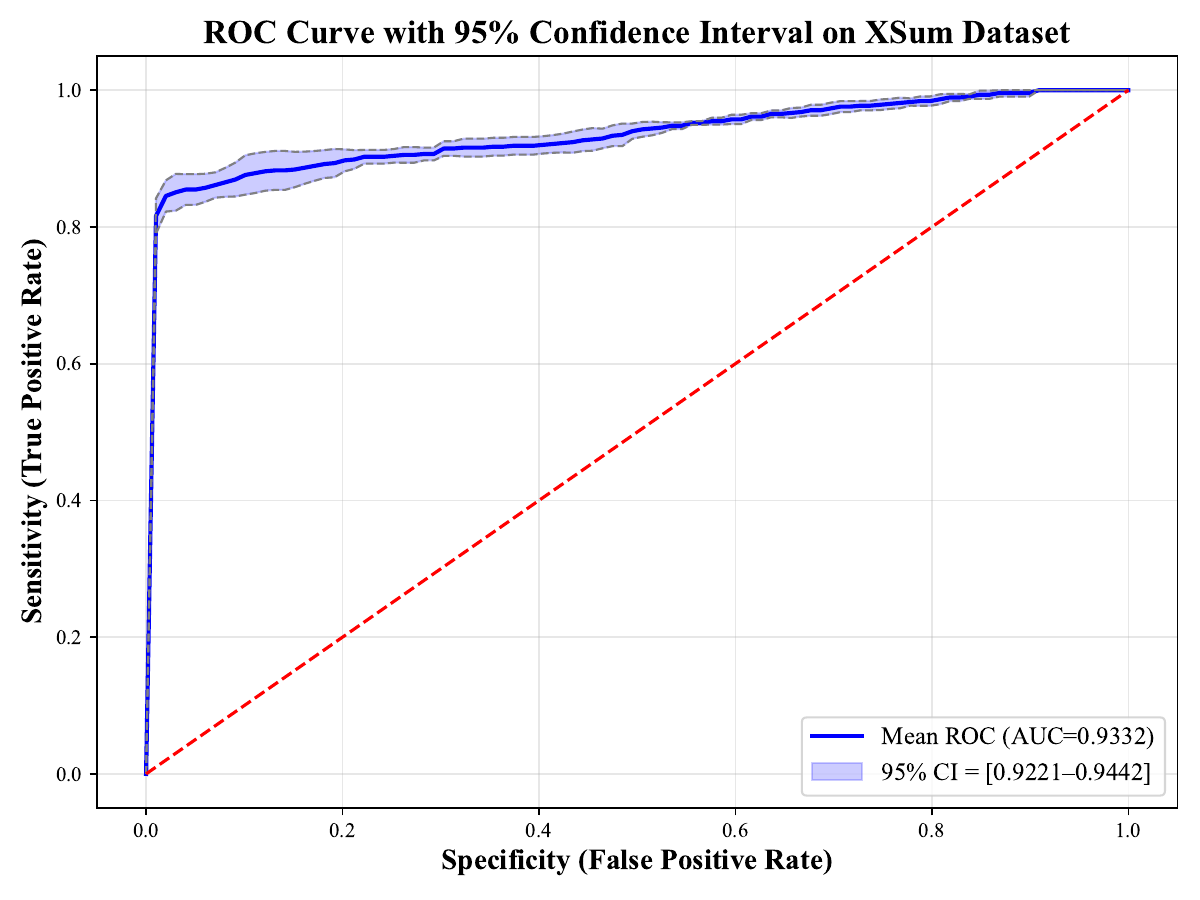}
    \hfill
    \includegraphics[width=0.48\linewidth,valign=t]{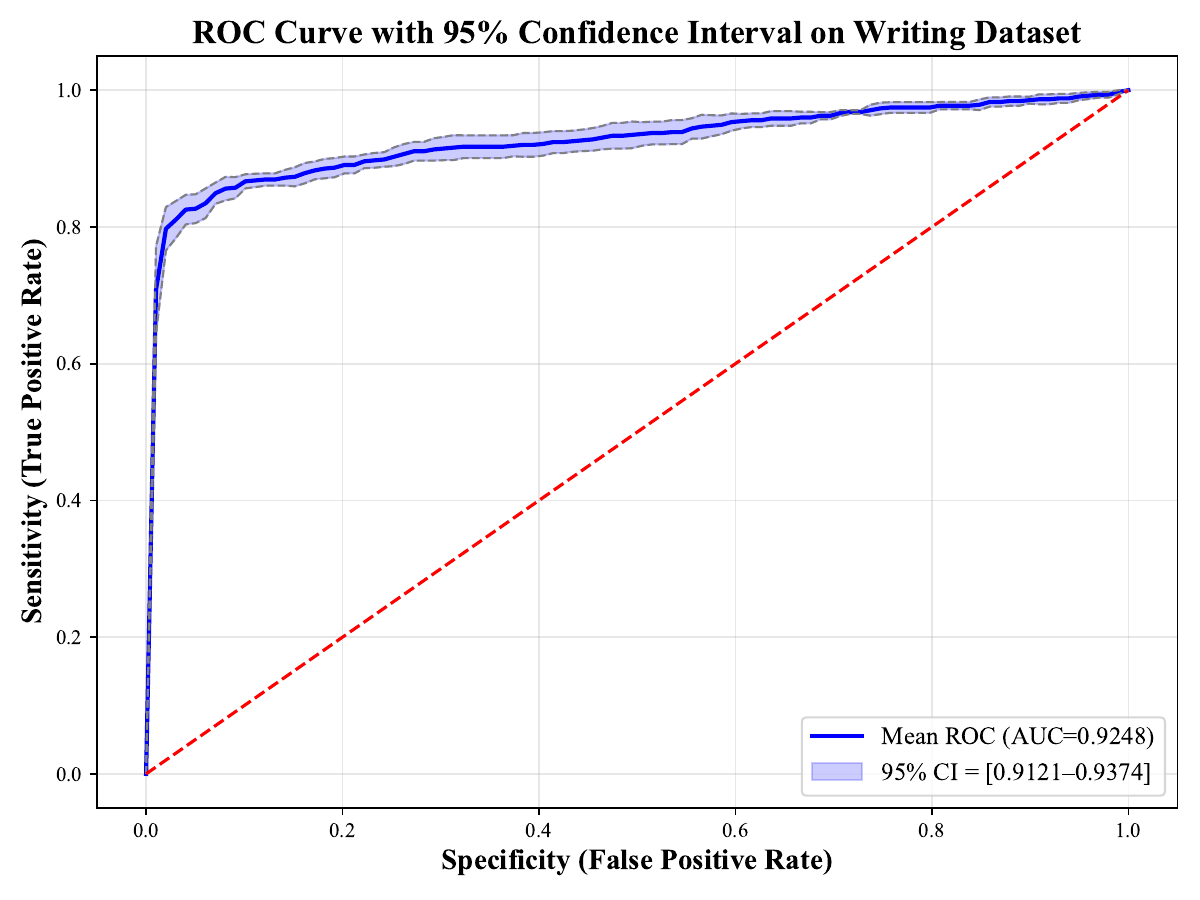}
    
    \vspace{2mm}
    
    \includegraphics[width=0.48\linewidth,valign=t]{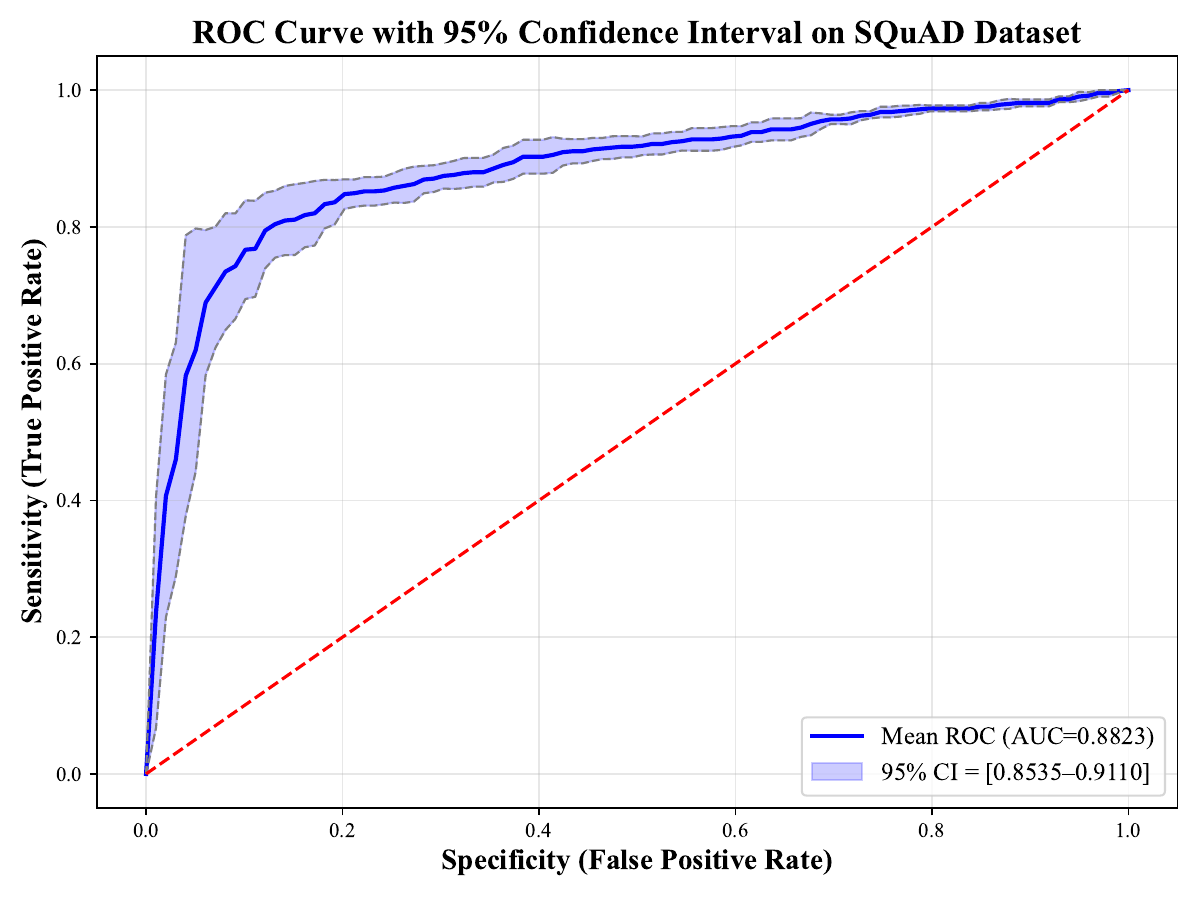}
    \hfill
    \includegraphics[width=0.48\linewidth,valign=t]{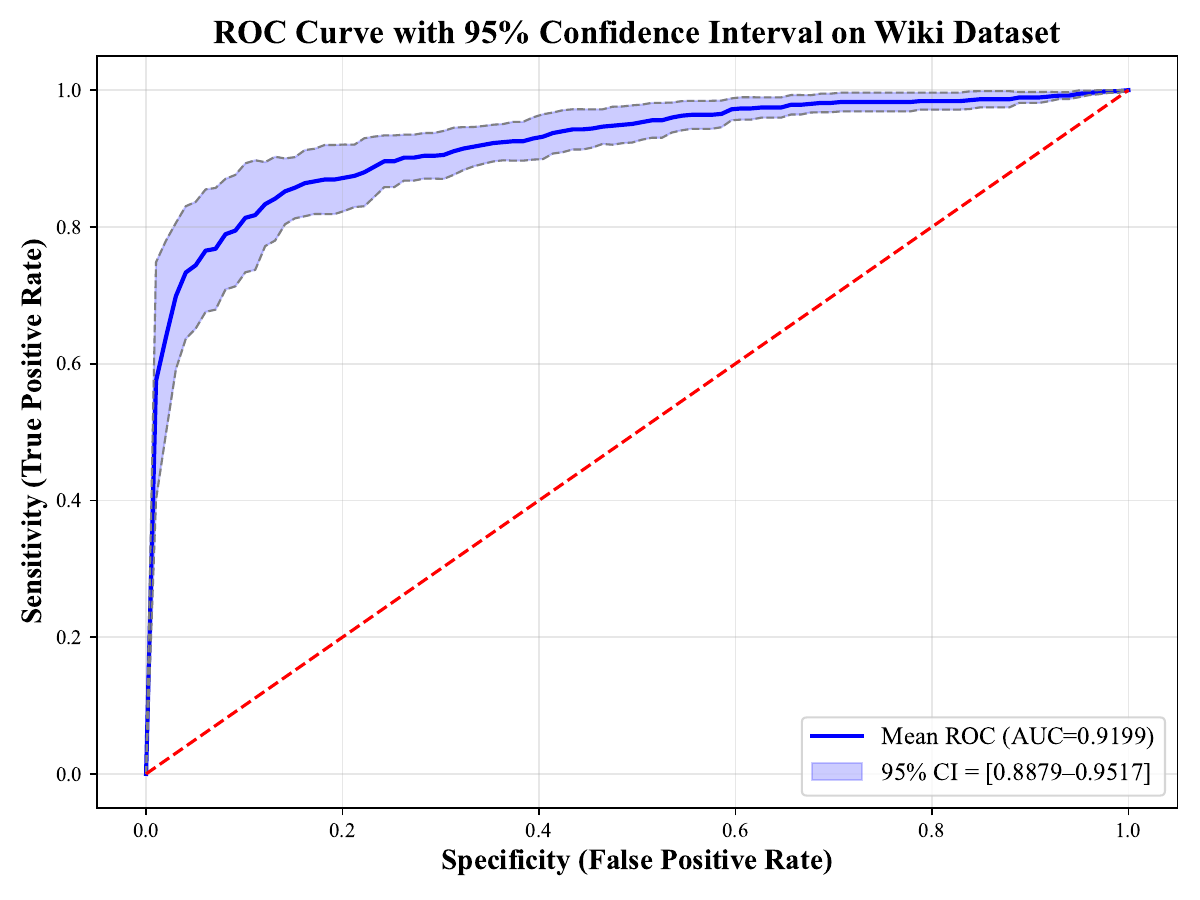}
    
    \caption{\textbf{ROC Curve with 95\% CI accross different datasets.}}
    \label{fig:roc-4grid}
\end{figure}

 \begin{figure}[h]
    \centering
    \includegraphics[width=0.75\linewidth,valign=t]{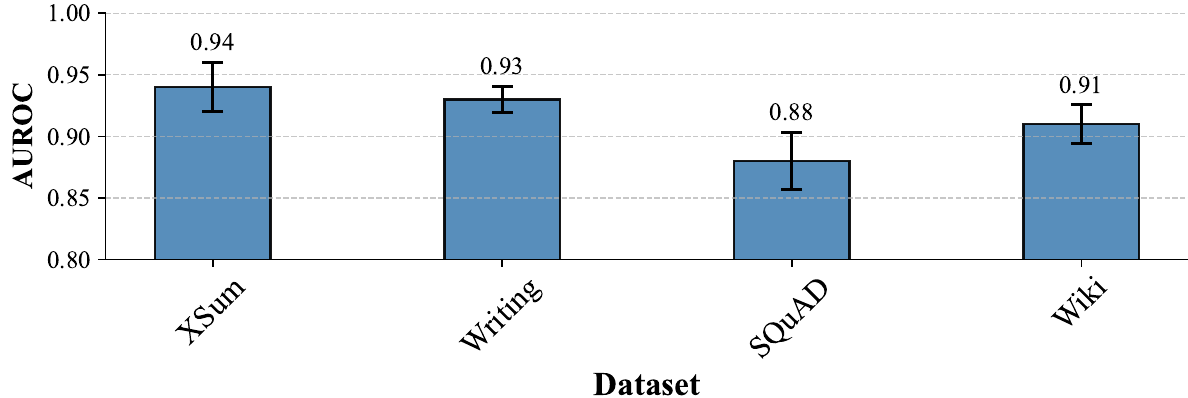}
    \caption{\textbf{Average AUROC with 95\% CI across different datasets.} }

\end{figure}

\newpage
\subsection{Source LLM Spec Details}
The specific versions of the LLM APIs used to construct the  datasets are detailed below:
\begin{itemize}
    \item OpenAI / GPT-3.5-turbo : \texttt{gpt-3.5-turbo-0125}
    \item OpenAI / GPT-4o : \texttt{gpt-4o-2024-05-13}
    \item OpenAI / GPT-o3 : \texttt{gpt-o3-2025-04-16}
    \item OpenAI / GPT-4 Turbo : \texttt{gpt-4-turbo-2024-04-09}
    \item Anthropic / Claude-3.5 : \texttt{claude-3-5-sonnet-2024-10-22}
    \item Google / Gemini-2.5Pro : \texttt{gemini-2.5-pro-exp-03-25}
    \item x.ai / Grok3 : \texttt{grok-3-reasoner}
    \item DeepSeek-AI / DeepSeek-R1 : \texttt{deepseek-r1(thinking)}
\end{itemize}

\subsection{Scoring \& Perturbation \& Generation model details.} 
In our HLPD detection framework, the generation model, perturbation model, and scoring model each serve distinct roles in the process. The generation model refers to the system originally used to revise human-written text and produce the suspect passage under investigation (e.g., a model like GPT-3.5 or Anthropic’s Claude). To detect AI-generated revisions, we then employ a perturbation model to generate minor variants of the suspicious text, and a scoring model to assign log-probability scores to both the original passage and its perturbed versions.

In a white-box scenario (where the source generation model is known and accessible), one can simply reuse the source model as the scoring model (and sometimes even as the perturbation model) to ensure consistency with the text’s generation process. By contrast, in black-box settings (where the source model is unknown or inaccessible), surrogate models must be used for both perturbation and scoring; prior work has shown that using a model like GPT-Neo-2.7B as a surrogate can be particularly effective in this regime. For example, DetectGPT perturbs text with a T5-3B model and uses the source model for scoring in a white-box setup, but in a black-box setting it replaces the scorer with GPT-Neo-2.7B. Similarly, Fast-DetectGPT introduces a conditional sampling strategy and relies on the source model for both perturbation and scoring under white-box conditions, whereas in black-box mode it employs GPT-J for generating perturbations and GPT-Neo-2.7B for scoring.

Given that our HLPD framework must detect revisions originating from multiple unknown source models, following ImBD, we adopt the black-box approach and use the fine-tuned GPT-Neo-2.7B model for both perturbation and scoring—essentially mirroring Fast-DetectGPT’s unified model approach in the white-box setting, but here using GPT-Neo-2.7B as a stand-in surrogate for any unknown source model.

\subsection{Detection metrics.} We mainly report the area under the receiver operating characteristic curve (AUROC) as a threshold-free evaluation metric. The ROC curve is a graph plotting TPR against the false positive rate = FP / (FP+TN) by varying a threshold. 

\subsection{Training Details}
\label{app:training details}
We fine-tune the gpt-neo-2.7B model from EleutherAI as our scoring model. All experiments are conducted on an Ubuntu 20.04 platform using a L20 (48GB) GPU, with Python 3.8, PyTorch 1.10.0, Cuda 12.1, Transformers 4.27.2, and Datasets 2.12.0. To ensure a fair comparison, the hyperparameters during the training process were the same as ImBD: 2 epochs, a learning rate of 0.0001.Each epoch takes 709 seconds on an L20 GPU.
 \subsection{Baseline Implementation Details}
 \begin{itemize}
\item Log-likelihood, Rank. These methods use an LLM to measure the token-wise log-probability and rank of words, then average the metric of each token to generate a score for the text. For the baseline experiments, we utilized GPT-neo-2.7B as their base model.

\item LRR. LRR used the Log-likelihood log-rank Ratio, which merges the benefits of log-likelihood and log-rank. We utilized GPT-neo-2.7B as their base model.

\item DetectGPT, NPR. DetectGPT and NPR are designed to measure changes in a model’s log-probability and log-rank function when slight perturbations are introduced to the original text. For the baseline experiments, we utilized GPT-neo-2.7B as their base model and T5-3B for paraphrasing, and we perturbed the text 100 times for each paragraph.

\item Fast-DetectGPT. Fast-DetectGPT shares the same spirit as DetectGPT but improves efficiency by using the conditional probability function derived from sampling with the base model, instead of relying on a separate perturbation model like T5. Following the original paper setting, we used GPT-J as a base model and GPT-neo-2.7B as a scoring model.

\item Ghostbuster. We deployed this method in complete alignment with its official GitHub repository. It analyzes text by utilizing a suite of weaker language models, like N-grams and early GPT-3 versions. It performs a structured search on their outputs to identify the most predictive features, which then train a classifier to detect AI-generated content. Crucially, same as hlpd (our method), it does not require access to the internal data (e.g., token probabilities) of the target model, making it highly effective at detecting text from "black-box" systems.

\item ReMoDetect,ImBD. Both methods are based on the principle of fine-tuning a model to have a stronger preference for machine-generated text, which amplifies the statistical differences for detection. Specifically, ImBD is designed to detect machine-revised text and fine-tunes the GPT-neo-2.7B model from EleutherAI as its scoring model, while ReMoDetect focuses on texts from aligned LLMs and utilizes the OpenAssistant reward model, which is based on DeBERTa-v3-Large.
 
\end{itemize}
 
\section{Framework for Machine Revision and Machine Generation}

\subsection{Generate task for machine-generated text }
\label{setting}
For generating purely machine-generated text, the large language model (LLM) was instructed to continue writing from a given prefix, without any other stylistic or restrictive requirements.\\ 
\begin{tcolorbox}[enhanced,breakable,title=Prompt Template:]
     \texttt{"You are a News writer. Please write an article with about 150 words starting exactly with: <prefix>" \# The \texttt{<prefix>} consisted of the first 30 tokens of a human-written sentence.}
\end{tcolorbox}

\subsection{Single task revised text generation }\label{single framework}
This section summarizes the data generation pipelines for single-task revisions as described by ImBD, which form a basis for comparison with our multi-task adversarial approach.

\noindent\textbf{Rewrite Task.}
The rewrite task  instructs the LLM  to act as a "professional rewriting expert" and paraphrase the provided human-written text  without losing details, while maintaining a similar length.
\begin{tcolorbox}[enhanced,breakable,title=Prompt Template:]
    \texttt{"You are a professional rewriting expert and you can help paraphrasing this paragraph  without  missing the original details. Please keep the length  of the rewritten text similar to the original text. <original human-written-text>"}
\end{tcolorbox}

\noindent\textbf{Polish Task.}
For the polish task, the LLM is asked to polish the supplied text under a randomly chosen style and length.
\begin{tcolorbox}[enhanced,breakable,title=Prompt Template:]
            \texttt{word\_lens = [15,30,50]  styles = ["formal", "oral", "academic", "literary",  "critical", "narrative", "descriptive", "lyric",  "objective", "subjective"]  <word\_len> = random.choice(word\_lens)  <style> = random.choice(styles) "Write a prompt in <word\_len> words that says you want gpt’s help in polishing a paragraph in a <style> style, this prompt can only be <word\_len> words or less."}
        \end{tcolorbox}

\noindent\textbf{Expand Task.}
The expand task asks the LLM to expand the original text with a randomly chosen style parameter. 
\begin{tcolorbox}[enhanced,breakable,title=Prompt Template:]
        \texttt{‘Expand but not extend the paragraph in a <style> style. The  paragraph to be expanded:<original human-written text> ’}
\end{tcolorbox}

\subsection{Multi-Task Adversarial Revised text generation}
\label{framework}
We propose a two-step generation pipeline for  multi-task adversarial machine-revised text generation, aiming to more accurately simulate realistic user scenarios. Specifically, the pipeline first generates a randomized user prompt utilizing the Deepseek-R1 model. Subsequently, we combine this generated user prompt with human-authored texts to LLMS to produce adversarially machine-revised texts.

\begin{tcolorbox}[enhanced,breakable,title=Prompt Template:]
   \texttt{REVISION\_GOALS = ["Paraphrase", "Rewrite", "Polish", "Expand", "Restructure"]\\
STYLE\_CONTROLS = ["formal", "oral", "academic", "literary", "critical",
                  "narrative", "descriptive", "lyric", "objective",
                  "subjective", "technical"]\\
ADVERSARIAL\_TEXT = [
    "make the paragraph sound as human as possible",
    "make this paragraph feel more natural, like a real person wrote it",
    "make this text sound less robotic and more human"
]\\
CONSTRAINTS = ["keep factual accuracy", "no hallucinated content", "maintain original intent"]\\
ADDITIONAL\_OPTS = ["enhance expression", "make sentences more concise", "reorganize the structure",
                   "preserve all factual details", "restructure sentences for better flow"]\\
word\_lens = [45, 50, 55]
word\_len = random.choice(word\_lens)
revision\_goal = random.choice(REVISION\_GOALS)\\
style = random.choice(STYLE\_CONTROLS)\\
adversarial = random.choice(ADVERSARIAL\_TEXT)\\
constraint = random.choice(CONSTRAINTS)\\
additional\_opt = random.choice(ADDITIONAL\_OPTS)\\
  "Create a prompt in <word\_len> words that says you want GPT's help to <revision\_goal> a paragraph in a <style> style, <adversarial>, <additional\_opt>, and <constraint}>. "

\end{tcolorbox}

The prompt template above illustrates multiple dimensions for generating diverse adversarial revision prompts. These include a set of revision goals ("Paraphrase," "Rewrite," "Polish," "Expand," "Restructure"), varying prompt lengths (45, 50, or 55 words), a variety of stylistic attributes (formal, oral, academic, literary, critical, narrative, descriptive, lyric, objective, subjective, technical), adversarial instructions aimed at enhancing the human-like quality of the text (e.g., "make the paragraph sound as human as possible," "make this text feel natural as if written by a real person," "make this text less robotic and more human"), and additional constraints and optional revision requests (such as "keep factual accuracy," "no hallucinated content," "maintain original intent," "enhance expression," "make sentences more concise," "preserve all factual details," and "restructure sentences for better flow"). From these predefined parameters, we randomly select elements from each dimension to systematically generate an initial prompt, which subsequently instructs the adversarial revision of the target text. The following are some real prompt examples generated by Deepseek-R1: \\ 


\begin{tcolorbox}[enhanced,breakable,title=Examples of prompts for multi-task adversarial revision.]
\emph{1. Polish this paragraph to sound as human as possible. Use a natural, conversational tone. Restructure sentences for better flow and clarity. Enhance with vivid, engaging descriptions while retaining all factual details. Ensure coherence and preserve original intent.}\\

\emph{2. Rewrite this paragraph to enhance human-like expression while preserving all factual details and original intent. Ensure clarity, factual accuracy, and avoid adding or omitting information. Prioritize varied sentence structure, relatable phrasing, and smooth transitions. Maintain a professional yet approachable tone. }\\ 

\emph{3. Please polish the provided paragraph to enhance expression, ensuring a natural, human tone. Maintain strict factual accuracy; avoid adding or altering information.  Restructure sentences for improved clarity, flow, and readability. Prioritize logical progression, smooth transitions, and avoid jargon. The final version should sound authentically human while retaining all essential information accurately.}
\end{tcolorbox}
\section{Humanization Details}

\subsection{Humanization Process flow diagram.}

We generate perturbed versions of the original machine-revised text and then use the HLPD model to select, 
fool the online detection platform GPTZero, making it think it was written by a human.
\begin{figure}[h]
\setlength{\abovecaptionskip}{0pt}   
\setlength{\belowcaptionskip}{0pt}    
\setlength{\floatsep}{0pt}            
\setlength{\textfloatsep}{0pt}
	\centering
		\centering
         \vspace{0pt}
		\includegraphics
            [width=1
		\textwidth]{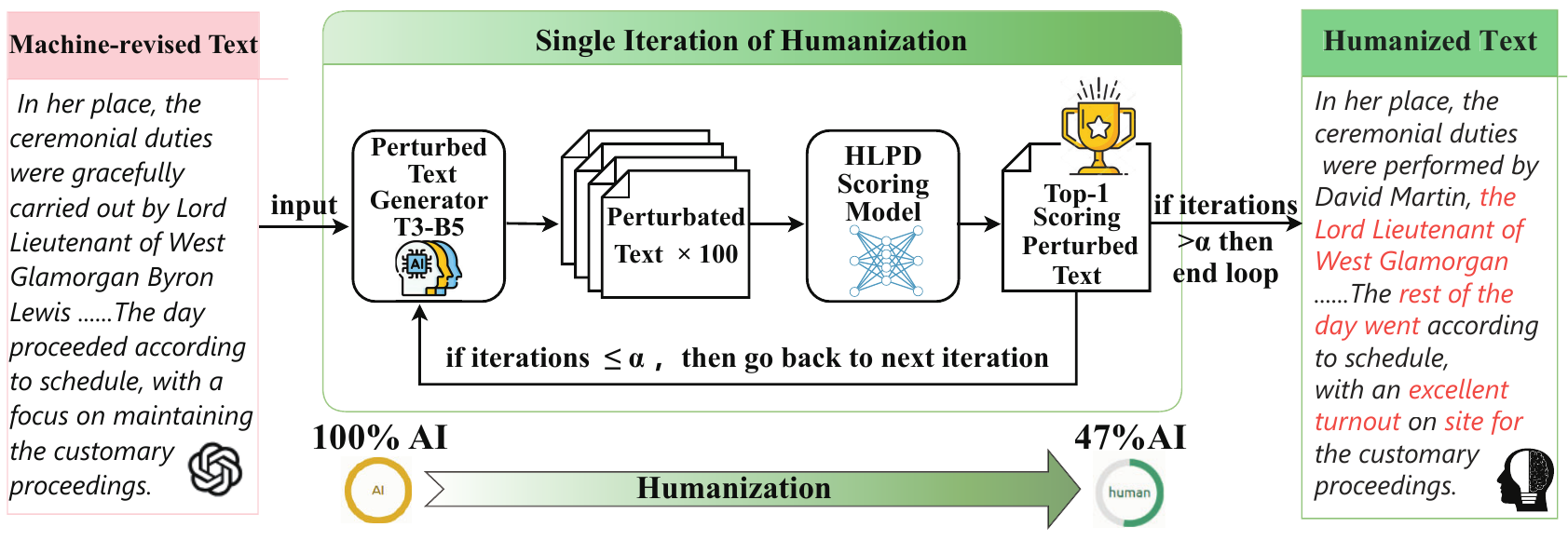}
        \vspace{0pt}
  	     \caption{ \textbf{Humanization of Machine-Revised Texts.} 
         }
        \label{fig:humanize}
         \vspace{0pt}
\end{figure}

\subsection{Additional Humanizing results }

\noindent\textbf{Humanizing against GPTZero.} As illustrated in Figure~\ref{fig:humanize-perf}, our iterative humanization technique proves highly effective at evading AI detection. Each iteration reduces the detection AUROC by an average of 5\%, culminating in a 20\% total reduction after four cycles compared to the original machine-revised text. This corresponds to a remarkable 74\% decrease in the AI probability score assigned by the commercial GPTZero platform, underscoring the practical efficacy of our method against contemporary detection systems.\\

\begin{figure}[h]
  \centering
  \includegraphics[width=0.98\linewidth]{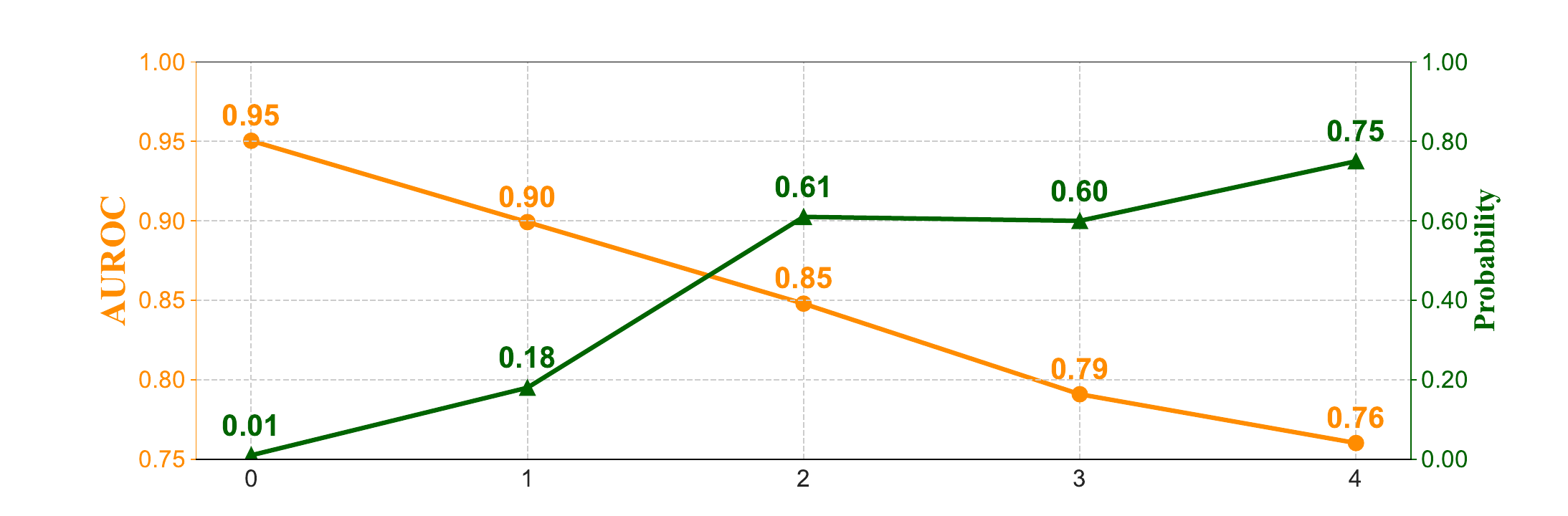}
  \caption{\textbf{Humanization performance on GPTZero platform.} Iterations denotes $x$ times humanization on machine-revised texts and 0 denotes the original machine-revised text.}
  \label{fig:humanize-perf}
\end{figure}

\label{app:imbd humanize}
\noindent\textbf{Humanizing against Baseline Web Demonstrations.} To further validate our method, we tested its evasive capabilities against the public web demonstrations for ImBD and Ghostbuster. We used the same sample texts as in the GPTZero evaluation, randomly selecting subsets of paragraphs to meet the 10,000-character limit imposed by the ImBD demo.The ImBD demo website link is: \textsf{\url{https://ai-detector.fenz.ai/ai-detector}}; the Ghostbuster demo website link is: \textsf{\url{https://ghostbuster.app/}.}

 Our humanization process proved highly effective. On the ImBD platform, the average AI detection probability fell progressively with each of the four iterations, dropping from an initial 81.83\% to 37.32\%—an average reduction of 11 percentage points per iteration. The effect was even more pronounced against Ghostbuster, where the AI probability score plummeted by 61 percentage points, from 87\% to 26\%. A qualitative example of a text before and after this transformation is presented below.\\

\subsection{Example of humanized Machine-revised text}
\begin{tcolorbox}[enhanced,breakable,title=Original machine-revised text.]
\emph{``John McCormack's reckless actions not only endangered the lives of innocent motorists but also shattered the peace and safety of a family. The terror and helplessness experienced by those on the road that day is unimaginable. The sound of metal colliding, the screech of tires, the moment of impact - all forever etched in their memories. The consequences of McCormack's dangerous driving will haunt them for years to come. The image of a lorry barreling towards them, a split-second decision to veer off the road, the relief of narrowly escaping tragedy - emotions that will linger long after the physical scars have healed. The sentence handed down, while just, cannot erase the trauma inflicted that day. The road to recovery for the family and those who witnessed the horror will be long and arduous. The impact of McCormack's actions reverberates far beyond the courtroom, leaving a trail of heartache and fear in its wake.''}
\end{tcolorbox}

\begin{tcolorbox}[enhanced,breakable,title=Machine-revised text after four times humanization.]
     \emph{``His actions not only endangered the lives of innocent motorists but, most dramatically, the life and safety of one family. The terror and fear felt by those travelling on the road that day will never be forgotten. The sound of an engine revving, the squeaking of brakes, the weight of fear on their faces - it will always be burned into their memories. The consequences of McCormack's dangerous driving will be seen and felt for years to come. The image of a car barrelling towards them, a split-second decision to veer off the road and the thrill of narrowly escaping tragedy are all images that will linger in their minds long after the physical scars have healed. The sentence handed down today is just, but doesn't reflect the magnitude of the tragedy that occurred that day. The road ahead for the Robinson family, and indeed all those who witnessed the event that day, will be long and painful - the impact of McCormack's actions reverberates far beyond the courtroom, leaving a trail of pain and suffering in its wake.''}
\end{tcolorbox}

\label{imp details}

 \section{Additional Experimental Results.}
\begin{table}[h]
\scriptsize
    \centering
    
\begin{threeparttable}
\setlength{\tabcolsep}{2pt} 

\begin{tabular}{l|c|cccccc|cccccc|c}
\toprule
\multirow{2}{*}{\textbf{Method}} &  \multirow{2}{*}{\textbf{Time cost}} & \multicolumn{5}{c}{\textbf{GPT-3.5}} &\multirow{2}{*}{ \textbf{ Avg.} }& \multicolumn{5}{c}{\textbf{GPT-4o}}&\multirow{2}{*}{ \textbf{ Avg.} }  & \multirow{2}{*}{ \textbf{\makecell[c]{Overall\\ Avg.} } }  \\
& & XSum & Writing & PubMed &SQuAD&Wiki&  & XSum & Writing & PubMed &SQuAD&Wiki&  \\
\midrule
\textbf{RoBERTa-base} & 0.07 & 0.5806 & 0.7225 & 0.4370 &0.6050 &0.5424 & 0.5775 & 0.4921 & 0.4774 & 0.2496 &0.3992 &0.4680 & 0.4172 & 0.4974 \\
\textbf{RoBERTa-large} & 0.11 & 0.6391 & 0.7236 & 0.4848 &0.6181 &0.6016 & 0.6134 & 0.4782 & 0.4708 & 0.3089 &0.3895 &0.4669 & 0.4235 & 0.5185 \\

\textbf{Likelihood} & 0.38 & 0.4982 & 0.8788 & 0.5528 &0.6521 & 0.7088& 0.6581 & 0.6436 & 0.8077 & 0.4596 &0.5384 & 0.6308& 0.6160 & 0.6371 \\
\textbf{Entropy} & 0.35 & 0.6742 & 0.3021 & 0.5662 &0.5927 & 0.4081& 0.5086 & 0.6122 & 0.2802 & 0.5899 & 0.5508& 0.4396& 0.4945 & 0.5016 \\
\textbf{LogRank} & 0.36 & 0.4711 & 0.8496 & 0.5597&0.6303 &0.6878 & 0.6397 & 0.4002 & 0.4936 & 0.4472 &0.5003 & 0.5920& 0.4867 & 0.5632 \\
\textbf{LRR} & 0.41 & 0.4016 & 0.7203 & 0.5629 &0.5472 & 0.5912& 0.5646 & 0.3095 & 0.4710 & 0.4710 &0.3798 & 0.4626& 0.4188 & 0.4917 \\
\textbf{DNA-GPT$\diamondsuit$} & 35.92 & 0.5338 & 0.8439 & 0.3333 &0.7083 &0.6056 & 0.6050 & 0.4974 & 0.7478 & 0.3151 &0.5873 &0.5759 & 0.5447 & 0.5748 \\
\textbf{NPR$\diamondsuit$} & 111.99 & 0.5659 & 0.8786 & 0.4246 & 0.6679& 0.6252& 0.6324 & 0.5065 & 0.8444 & 0.3740 &0.5371 &0.5859& 0.5696 & 0.6010 \\
\textbf{DetectGPT$\diamondsuit$} & 111.33 & 0.6343 & 0.5608 & 0.4949 &0.7029 & 0.6854& 0.6117 & 0.6217 & 0.8771 & 0.5612 &0.6458 &0.6991 & 0.6810 & 0.6463 \\ 
\textbf{Fast-Detect-GPT} & 0.72 & 0.7312 & 0.7182 & 0.4842 & 0.8506& 0.7620& 0.7092 & 0.6293 & 0.6175 & 0.6178 &0.7184 & 0.7257& 0.6637 & 0.6865 \\
\textbf{ImBD}  & 0.72 &0.9849 & 0.9871 & \textbf{0.8626} &0.9530 & 0.9486& 0.9472 & 0.9486&0.9468 & 0.7743   &0.8880&0.9400 & 0.8995 & 0.9234 \\
\rowcolor{gray!25}
\textbf{HLPD } &0.72&	\textbf{0.9923}&	\textbf{0.9748}&	0.8446&\textbf{0.9720}	&\textbf{0.9880} & \textbf{0.9623}&	\textbf{0.9804}&	\textbf{0.9658}&	\textbf{0.8968}&\textbf{0.9362}	& \textbf{0.9888}& \textbf{0.9456}& \textbf{0.9580}\\ \bottomrule
\end{tabular}

 \begin{tablenotes}
        \footnotesize
        \scriptsize
        \item[*]All mentions of the symbol $\diamondsuit$  in this paper denotes methods that call the model a hundred times, thereby resulting in a substantial increase in computational load. The metric for time cost is the number of seconds required to process 1,000 words.
        \item[**]The scoring model typically relies on Neo-2.7B  as the source. In contrast, both NPR and DetectGPT leverage T5-3B to create perturbations, while Fast-DetectGPT adopts GPT-J  as a surrogate model for generating samples.

      \end{tablenotes}
\end{threeparttable}
\caption{\textbf{Detection of GPT-3.5 and GPT-4o polished text.} Metric: AUROC.}
\label{tab:polished_text_detection}
\end{table}

\noindent\textbf{Detailed Results for Machine-Polished Text Detection.} As presented in Table~\ref{tab:polished_text_detection} and~\ref{gptzeroc} our method demonstrates superior performance in detecting texts revised by GPT-3.5-turbo and gpt-4o-2024-05-13. It consistently achieved the highest accuracy across all evaluated datasets—XSUM, SQuAD, PubMed, Wikitext and WritingPrompts—outperforming all competing methods.
\newpage
\begin{figure}[htbp!]

    \begin{minipage}[t]{0.58\linewidth}
    \footnotesize
        \flushleft 

        
       \setlength{\tabcolsep}{6pt} 
\begin{tabular}{llcccc} 
\toprule
\multirow{2}{*}{\textbf{Task}} 
  & \multirow{2}{*}{\textbf{Method}} 
  & \multicolumn{3}{c}{\textbf{Dataset}}&\multirow{2}{*}{\textbf{Avg.}}                      \\ 
\cmidrule(lr){3-5}
  & 
  & {\textbf{XSum}} & {\textbf{Writing}} & {\textbf{PubMed}} & \\
\midrule
\multirow{4}{*}{\textbf{Rewrite}}
  & \textbf{GPTZero}           & 0.7156 & \textbf{0.8756} & 0.7159            & 0.7690          \\ 
  & \textbf{ImBD}              & 0.7995 & 0.8136            & 0.6178            & 0.7436          \\
  & \cellcolor{gray!25}\textbf{HLPD} & \cellcolor{gray!25}\textbf{0.9619} & \cellcolor{gray!25}0.8435 & \cellcolor{gray!25}\textbf{0.7393} & \cellcolor{gray!25}\textbf{0.8481} \\
\midrule
\multirow{4}{*}{\textbf{Polish}}
  & \textbf{GPTZero}           & 0.9442 & \textbf{0.9841} & 0.8825            & 0.9369          \\ 
  & \textbf{ImBD}              & 0.9486 & 0.9468            & 0.7743            & 0.8899          \\
  & \cellcolor{gray!25}\textbf{HLPD} & \cellcolor{gray!25}\textbf{0.9804} & \cellcolor{gray!25}0.9658 & \cellcolor{gray!25}\textbf{0.8968} & \cellcolor{gray!25}\textbf{0.9477} \\
\bottomrule
\end{tabular}
\centering
\captionof{table}{\textbf{Comparison with GPTZero on detecting GPT-4o rewritten and polished text.} Metric: AUROC.}
        \label{gptzeroc}
        \vspace{2.5em} 

        
          \setlength{\tabcolsep}{8pt} 
        
        \begin{tabular}{lcccc}
            \toprule
            \multirow{2}{*}{\textbf{Method}} & \multicolumn{3}{c}{\textbf{Tasks}} & \multirow{2}{*}{\textbf{Avg.}} \\ 
             \cmidrule(lr){2-4}
            & \textbf{Rewrite} & \textbf{Expand} & \textbf{Polish} &   \\ \midrule
            \textbf{Likelihood}& 0.4073 & 0.4564 & 0.6039  & 0.4892 \\
            \textbf{Entropy} & 0.5840 & 0.6629 & 0.5431  & 0.5967 \\
            \textbf{LogRank} & 0.3868 & 0.4273 & 0.5864  &  0.4635\\
            \textbf{LRR} & 0.3488 & 0.3581 & 0.5183 &  0.4084 \\
            \textbf{DNA-GPT} & 0.4101 & 0.4901 & 0.5847  &0.4950\\
            \textbf{NPR} & 0.3606 & 0.5139 & 0.5673  & 0.4806 \\
            \textbf{DetectGPT} & 0.4060 & 0.6000 & 0.6615  &  0.5558 \\
            \textbf{Fast-DetectGPT} & 0.4499 & 0.7159 & 0.7989 &  0.6549 \\
            \textbf{ImBD} & 0.8739 & \textbf{0.9758} & 0.9707 & 0.9401\\
            \rowcolor{gray!25}
            \textbf{HLPD} & \textbf{0.9787} & 0.9679& \textbf{0.9895} &  \textbf{0.9787}\\
            \bottomrule
        \end{tabular}
        \centering
\captionof{table}{\textbf{Performance on diverse tasks.} Metric: AUROC.}
\label{tab:diverse_tasks}

    \end{minipage}%
    \hfill 
    \begin{minipage}[t]{0.4\linewidth}
        \raisebox{-0.77\height}{%
            \includegraphics[width=\linewidth]{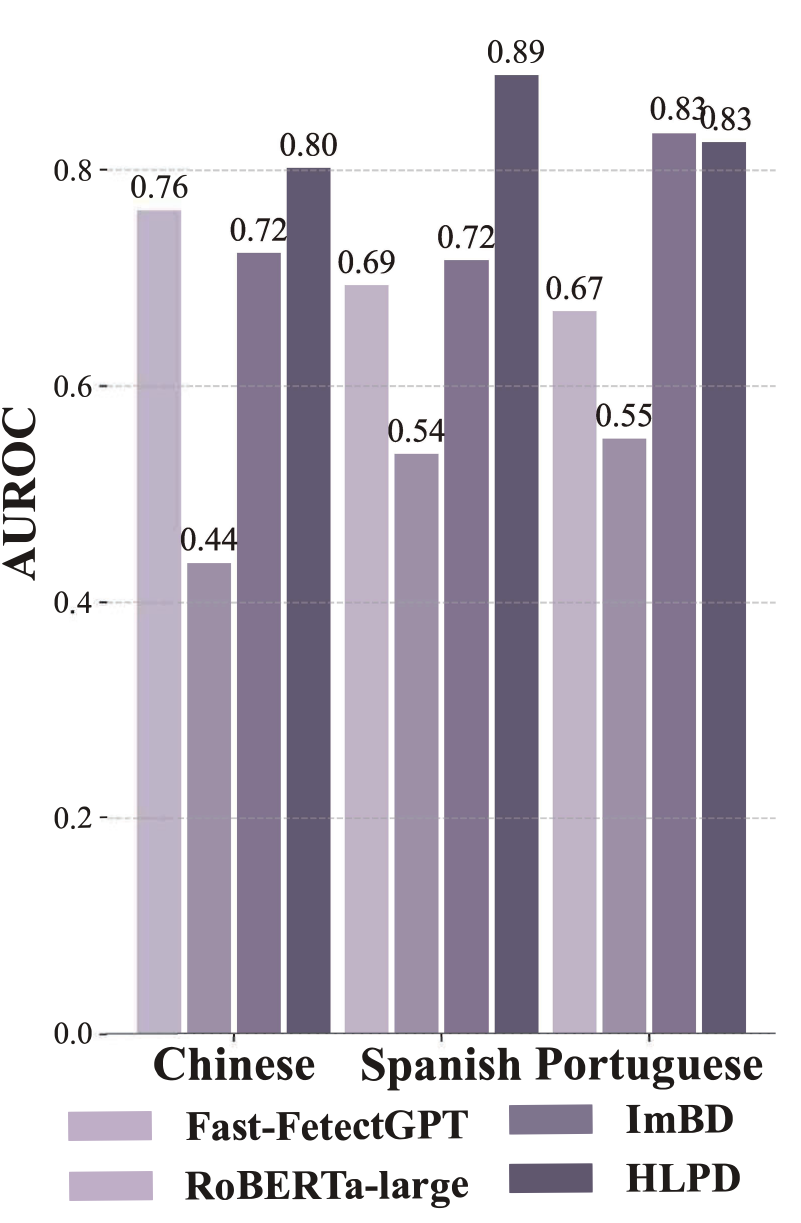}%
        }
        \captionof{figure}{\textbf{Multilingual Detection}}
        \label{pic:multilanguage}
    \end{minipage}

\end{figure}
\noindent\textbf{Detection Performance on diverse languages.} As shown in Figure~\ref{pic:multilanguage}, HLPD's performance was evaluated on Chinese, Portuguese, and Spanish texts polished by GPT-3.5. Across all languages, it consistently outperformed ImBD by 10\% and RoBERTa-large by an average of 33\%.\\

\noindent\textbf{Robustness to Diverse Revision Goals.} As detailed in Table~\ref{gptzeroc}, \ref{tab:diverse_tasks}, HLPD shows strong robustness against diverse revision goals. On the XSum dataset, revised by six different LLMs across three revision tasks, HLPD achieved the highest average AUROC. This represents a 32.38\% improvement over Fast-DetectGPT and a 4.11\% gain over ImBD, underscoring its strong multi-task detection capabilities.

\begin{table}[htbp!]
    \centering

    \setlength{\tabcolsep}{21pt} 
    \sisetup{table-format=1.4}

    \begin{tabular}{@{} l *{5}{S} @{}}
        \toprule
        
        \multirow{2}{*}{\textbf{Dataset}} & \multicolumn{5}{c}{\textbf{Iteration Revision}} \\
        \cmidrule(lr){2-6}
        
        & {\textbf{1}} & {\textbf{2}} & {\textbf{3}} & {\textbf{4}} & {\textbf{5}} \\
        \midrule
        \textbf{XSum}    & 0.9523 & 0.9611 & 0.9893 & 0.9703 & 0.9905 \\
        \addlinespace 
        \textbf{Writing} & 0.9446 & 0.9537 & 0.9750 & 0.9727 & 0.9860 \\
        \addlinespace
        \textbf{SQuAD}   & 0.9044 & 0.8958 & 0.9436 & 0.9269 & 0.9307 \\
        \addlinespace
        \textbf{Wikitext}    & 0.9483 & 0.9469 & 0.9790 & 0.9527 & 0.9711 \\
        \bottomrule
    \end{tabular}
    \caption{\textbf{Performance on Multi-Iterative Revision.} Metric: AUROC.}
    \label{tab:multi_iteration_enhanced}
\end{table}

\noindent\textbf{Robustness to Multiple Revision Iterations.} To assess its robustness to iterative revisions, we tested HLPD on texts revised one to five times by GPT-4o and Gemini-2.5-Pro. As shown in Table~\ref{tab:multi_iteration_enhanced}, HLPD consistently maintained the highest average AUROC scores, demonstrating its stable detection efficacy against extensive machine revisions.\\    

\noindent\textbf{Detailed results on open-source model revised text.} The results presented in Tables \ref{tab:detals on open source model polished} and \ref{tab:detals on open source model rewritten} underscore the clear superiority of our method for detecting content from open-source LLMs. This superior performance holds true regardless of the generative model, proving effective against Qwen2-7B, Llama-3-8B, Mistral-7B, and Deepseek-7B.\\  

\noindent\textbf{Detailed results on diverse revision tasks and target LLMs.} Table~\ref{tab:performance across different Models and Tasks.} demonstrates HLPD's superior performance across various tasks, including text generation, rewriting, polishing, and expansion. Our method consistently outperformed baselines against all tested LLMs. A key advantage is its black-box compatibility: HLPD effectively detects outputs from various models without requiring access to their internal data (e.g., token probabilities). This highlights its strong generalization and makes it highly practical for real-world applications.

\begin{figure}[h]
  \centering
  \includegraphics[width=0.98\linewidth]{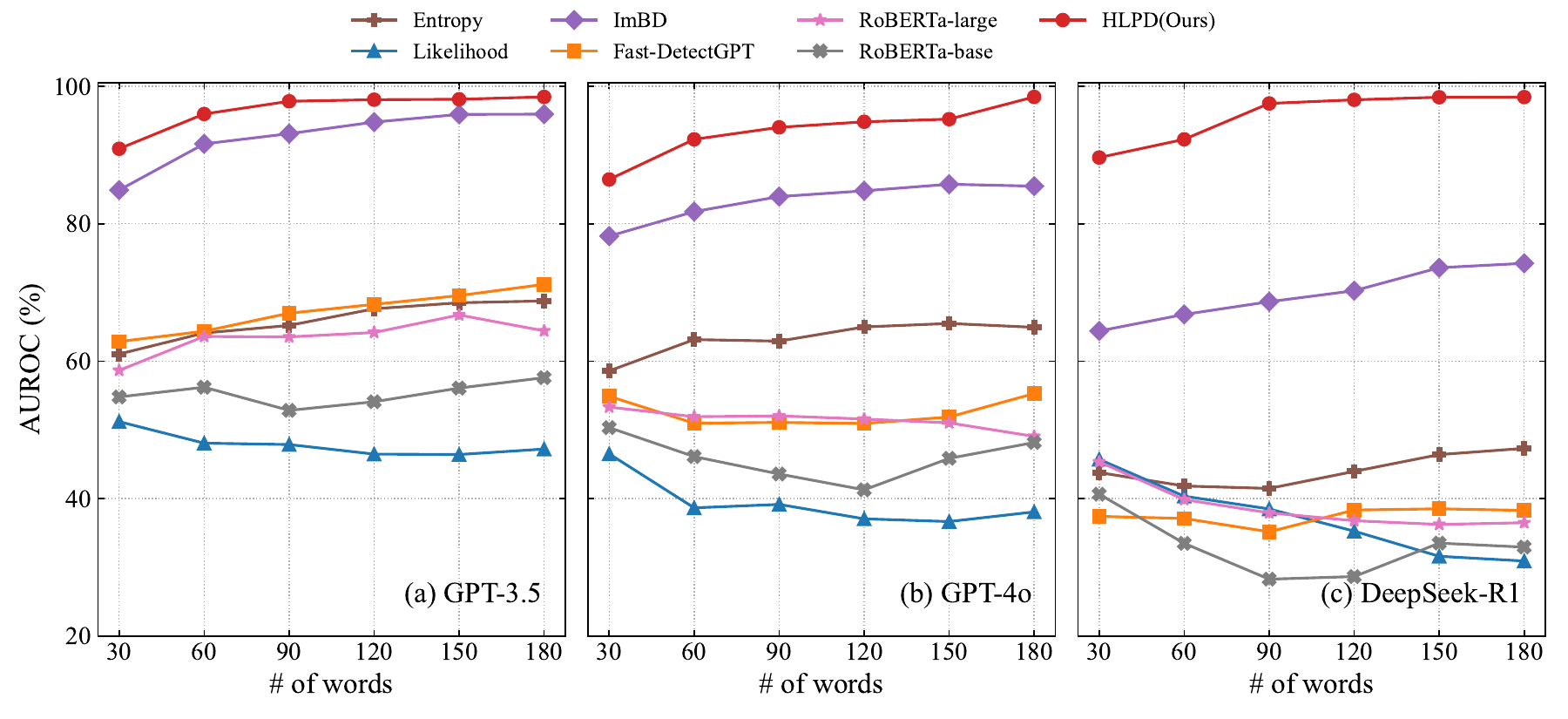}
  \caption{\textbf{Detection on different passage length for polish task.} }
  \label{fig:humanize-perf11}
\end{figure}

\begin{figure}[h]
  \centering
  \includegraphics[width=0.98\linewidth]{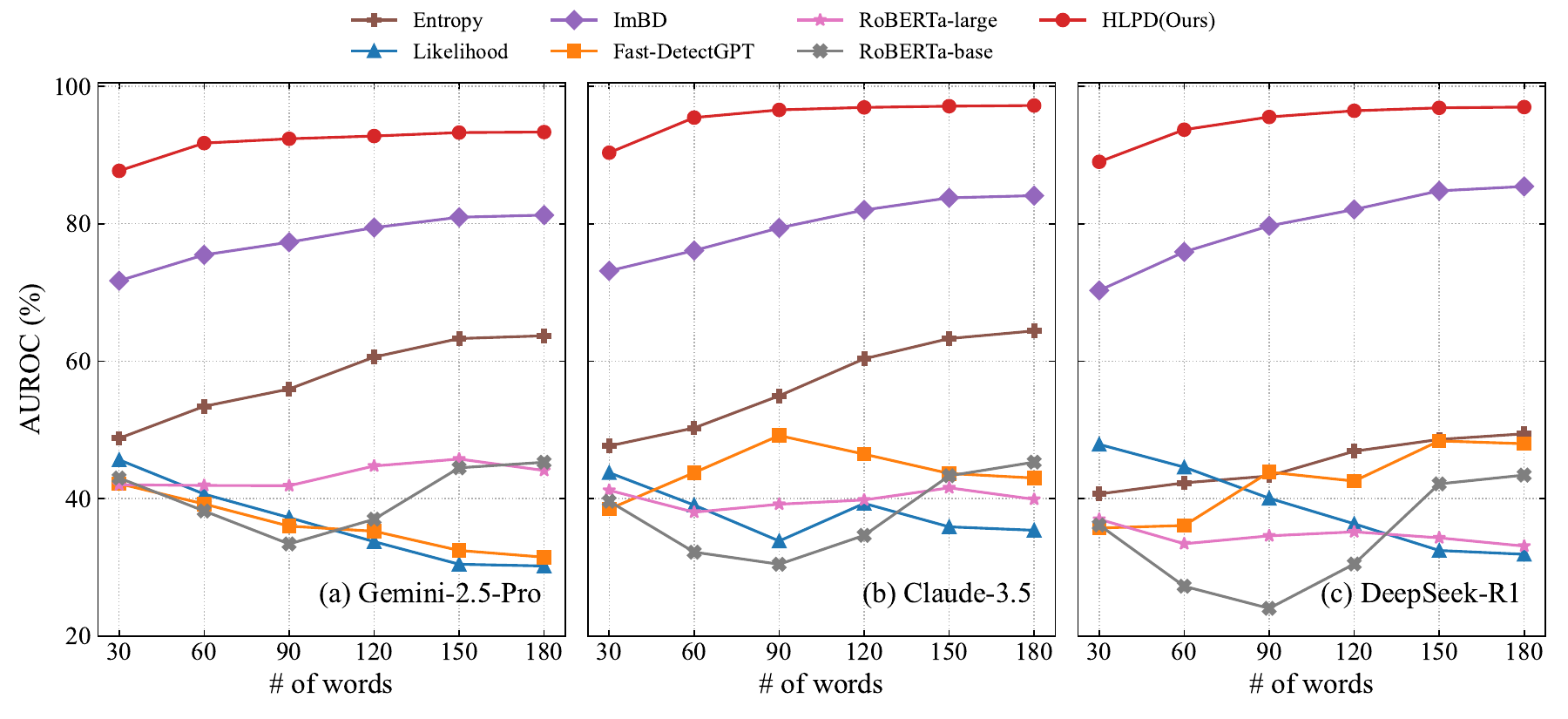}
  \caption{\textbf{Detection on different passage length for adversarial multi-task revision.} }
  \label{fig:humanize-perf22}
\end{figure}

\noindent\textbf{Results on different passage length.} To evaluate its robustness against varying text lengths, we tested HLPD on passages of different lengths, using truncations by Deepseek-R1, GPT-3.5, and GPT-4o for the polish task, and by Deepseek-R1, Claude-3.5, and Gemini-2.5-Pro for the adversarial multi-task revision task. As shown in Figures \ref{fig:humanize-perf11} and \ref{fig:humanize-perf22}, HLPD consistently achieved the highest average AUROC scores, demonstrating its superior performance and stability, even in sentence-level detection, across different passage lengths.

\begin{table}[h]
\centering

\resizebox{\textwidth}{!}{
\begin{tabular}{l l c c c c c}
\toprule
 \multirow{2}{*}{\textbf{Dataset}} & \multirow{2}{*}{\textbf{Method}} &\multicolumn{4}{c}{\textbf{SourceModel}}& \multirow{2}{*}{\textbf{Avg.}}\\
 & & \textbf{Qwen2-7B} & \textbf{Llama-3-8B} & \textbf{Mistral-7B} & \textbf{Deepseek-7B} & \\
\midrule
\multirow{8}{*}{\textbf{XSum}}
 & Likelihood 
 & 0.2520 & 0.5695 & 0.4335 & 0.5438 & 0.4502 \\
 & Entropy 
 & 0.7623 & 0.6348 & 0.6539 & 0.6402 & 0.6728 \\
 & LogRank 
 & 0.2246 & 0.5412 & 0.3980 & 0.5288 & 0.4232 \\
 & LRR 
 & 0.1875 & 0.4530 & 0.3112 & 0.4859 & 0.3594 \\
 & NPR 
 $\diamondsuit$       & 0.3896 & 0.6144 & 0.4594 & 0.5476 & 0.5028 \\
 & DetectGPT 
 $\diamondsuit$ & 0.4885 & 0.6904 & 0.5480 & 0.6172 & 0.5860 \\
 & Fast-DetectGPT 
 & 0.5945 & 0.8192 & 0.7034 & 0.8177 & 0.7337 \\
   &ImBD
   &0.9589 &\textbf{0.9884} &0.9671&0.9764&0.9727\\
    \rowcolor{gray!25}
 & HLPD                         & \textbf{0.9951} &0.9841& \textbf{0.9947}  & \textbf{0.9902} & \textbf{0.9910} \\

\midrule
\multirow{8}{*}{\textbf{SQuAD}}
 & Likelihood   & 0.3635 & 0.6388 & 0.5633 & 0.6408 & 0.5516 \\
 & Entropy      & 0.6931 & 0.5920 & 0.5935 & 0.5426 & 0.6068 \\
 & LogRank      & 0.3395 & 0.6410 & 0.5368 & 0.6167 & 0.5268 \\
 & LRR                & 0.2296 & 0.5150 & 0.4524 & 0.5291 & 0.4490 \\
 & NPR $\diamondsuit$       & 0.4399 & 0.6511 & 0.5479 & 0.5449 & 0.5460 \\
 & DetectGPT $\diamondsuit$ & 0.5396 & 0.7223 & 0.6410 & 0.6320 & 0.6339 \\
 & Fast-DetectGPT    & 0.7056 & 0.8855 & 0.8317 & 0.8344 & 0.8143 \\
 &ImBD &0.8860 &0.9508 &0.9136 &0.9161 &0.9166\\
  \rowcolor{gray!25}
 & HLPD                         & \textbf{0.9801} & \textbf{0.9730} & \textbf{0.9833} & \textbf{0.9513} & \textbf{0.9719} \\

\midrule
\multirow{8}{*}{\textbf{WritingPrompts}}
 & Likelihood  & 0.4354 & 0.8435 & 0.5133 & 0.7708 & 0.6408 \\
 & Entropy      & 0.6013 & 0.4342 & 0.5140 & 0.3579 & 0.4769 \\
 & LogRank      & 0.3810 & 0.8068 & 0.4640 & 0.7366 & 0.5996 \\
 & LRR               & 0.2945 & 0.6148 & 0.3282 & 0.6494 & 0.4667 \\
 & NPR $\diamondsuit$       & 0.3684 & 0.8101 & 0.5309 & 0.7240 & 0.6199 \\
 & DetectGPT $\diamondsuit$ & 0.6323 & 0.8380 & 0.5877 & 0.7518 & 0.7025 \\
 & Fast-DetectGPT    & 0.8967 & 0.9562 & 0.9141 & 0.9539 & 0.9302 \\
  & ImBD &0.9653 &\textbf{0.9908} &0.9670 &\textbf{0.9796} &0.9757\\
 \rowcolor{gray!25}
 & HLPD                         & \textbf{0.9980} & 0.9724 & \textbf{0.9928} & 0.9614 & \textbf{0.9812} \\

\midrule
\multirow{8}{*}{\textbf{PubMed}}
 & Likelihood   & 0.4367 & 0.5900 & 0.5951 & 0.7119 & 0.6084 \\
 & Entropy      & 0.6846 & 0.6062 & 0.6390 & 0.5274 & 0.6153 \\
 & LogRank      & 0.4346 & 0.5969 & 0.5900 & 0.7174 & 0.5847 \\
 & LRR               & 0.4362 & 0.5945 & 0.5608 & 0.6980 & 0.5724 \\
 & NPR $\diamondsuit$       & 0.3117 & 0.4052 & 0.3848 & 0.4269 & 0.4572 \\
 & DetectGPT $\diamondsuit$ & 0.6122 & 0.7660 & 0.6639 & 0.7024 & 0.6861 \\
 & Fast-DetectGPT    & 0.7347 & 0.8892 & 0.8685 & 0.8875 & 0.8450 \\
   & ImBD     & 0.8545 & 0.8823
 &0.9482 &0.9466 & 0.9079 \\
 \rowcolor{gray!25}
 & HLPD                         & \textbf{0.9933} & \textbf{0.9643} & \textbf{0.9809} & \textbf{0.9860} & \textbf{0.9811} \\
\midrule
\multirow{8}{*}{\textbf{Wikitext}}
 & Likelihood   & 0.5408 & 0.7848 & 0.7166 & 0.7571 & 0.6998 \\
 & Entropy      & 0.5069 & 0.3555 & 0.4065 & 0.3764 & 0.4113 \\
 & LogRank      & 0.5173 & 0.7662 & 0.6964 & 0.7390 & 0.6797 \\
 & LRR               & 0.4644 & 0.6745 & 0.6091 & 0.6429 & 0.5977 \\

 & NPR $\diamondsuit$       & 0.5505
 & 0.6961 & 0.6283 & 0.6088 & 0.6209 \\
 & DetectGPT $\diamondsuit$ & 0.6545 & 0.7662 & 0.7117 & 0.6931 & 0.7064\\
 & Fast-DetectGPT    & 0.6841 & 0.8728 & 0.6480 & 0.8036 & 0.7521\\
   & ImBD     & 0.8973 & 0.9304
 & 0.9128 &0.9263 &0.9167  \\
 \rowcolor{gray!25}
 & HLPD                         & \textbf{0.9940} &\textbf{0.9643} & \textbf{0.9902} & \textbf{0.9860} & \textbf{0.9836} \\
\bottomrule
\end{tabular}
}
\caption{\textbf{Performance on \textit{open-source} model polished text.} Metric: AUROC. AUROC scores are averaged across the datasets generated by the polish task based on \textit{XSum}, \textit{SQuAD}, \textit{PubMed}, \textit{Wikitext} and \textit{WritingPrompts}. The scoring model typically relies on Neo-2.7B  as the source. In contrast, both NPR and DetectGPT leverage T5-3B  to create perturbations, while Fast-DetectGPT adopts GPT-J  as a surrogate model for generating samples.}
\label{tab:detals on open source model polished}
\end{table}

\begin{table}[htbp]
\footnotesize
\centering

\resizebox{\textwidth}{!}{%
\begin{tabular}{l l c c c c c}
\toprule
\multirow{2}{*}{\textbf{Dataset}} & \multirow{2}{*}{\textbf{Method}} &\multicolumn{4}{c}{\textbf{SourceModel}}& \multirow{2}{*}{\textbf{Avg.}}\\
 & & \textbf{Qwen2-7B} & \textbf{Llama-3-8B} & \textbf{Mistral-7B} & \textbf{Deepseek-7B} & \\\midrule
\multirow{10}{*}{\textbf{XSum}} 
  & Likelihood     & 0.2741 & 0.5851 & 0.3613 & 0.5170 & 0.4344 \\
  & Entropy        & 0.6396 & 0.5165 & 0.6028 & 0.5862 & 0.5863 \\
  & LogRank       & 0.2564 & 0.5589 & 0.3399 & 0.5053 & 0.4151 \\
  & LRR                 & 0.2376 & 0.4905 & 0.3071 & 0.4742 & 0.3774 \\
 
  & NPR  $\diamondsuit$           & 0.2443 & 0.4986 & 0.2888 & 0.4380 & 0.3674 \\
  & DetectGPT  $\diamondsuit$  & 0.2726 & 0.5436 & 0.3115 & 0.4512 & 0.3947 \\
  & Fast-DetectGPT      & 0.2853 & 0.6911 & 0.3938 & 0.6647 & 0.5087 \\
    & ImBD    &0.8952 &0.9710 &0.8348 &0.8739&0.8937\\
   \rowcolor{gray!25}
 & HLPD                          & \textbf{0.9988} & \textbf{0.9966} & \textbf{0.9940} & \textbf{0.9212} & \textbf{0.9777} \\

\midrule
\multirow{9}{*}{\textbf{SQuAD}}
  & Likelihood    & 0.3657 & 0.6584 & 0.5017 & 0.6540 & 0.5450 \\
  & Entropy      & 0.5718 & 0.4639 & 0.5128 & 0.4514 & 0.5000 \\
  & LogRank       & 0.3401 & 0.6380 & 0.4843 & 0.6432 & 0.5264 \\
  & LRR                 & 0.2925 & 0.5385 & 0.4256 & 0.5858 & 0.4606 \\
  & NPR  $\diamondsuit$             & 0.3238 & 0.5539 & 0.4346 & 0.5490 & 0.4653 \\
  & DetectGPT  $\diamondsuit$  & 0.3550 & 0.6051 & 0.4609 & 0.5763 & 0.4993 \\
  & Fast-DetectGPT     & 0.3764 & 0.7425 & 0.5272 & 0.7313 & 0.5944 \\
  & ImBD     &  0.7874 &0.9089 &0.7683 &0.7716 &0.8091  \\
  \rowcolor{gray!25}
  & HLPD                          & \textbf{0.9884} & \textbf{0.9811} & \textbf{0.9524} & \textbf{0.8152} & \textbf{0.9343} \\

\midrule
\multirow{9}{*}{\textbf{WritingPrompts}}
  
  & Likelihood    & 0.4354 & 0.8435 & 0.5133 & 0.7708 & 0.6408 \\
  & Entropy       & 0.6013 & 0.4342 & 0.3442 & 0.5440 & 0.4710 \\
  & LogRank       & 0.3810 & 0.8068 & 0.4640 & 0.7466 & 0.5996 \\
  & LRR                 & 0.2457 & 0.6148 & 0.3282 & 0.6494 & 0.4647 \\
  & NPR  $\diamondsuit$             & 0.3684 & 0.8101 & 0.5309 & 0.7240 & 0.6184 \\
  & DetectGPT  $\diamondsuit$  & 0.6323 & 0.8380 & 0.5877 & 0.7518 & 0.7025 \\
  & Fast-DetectGPT     & 0.6089 & 0.9338 & 0.6480 & 0.8408 & 0.7579 \\
   & ImBD     &  0.8845 &0.9761 &0.8384 &\textbf{0.9020} &0.9003  \\
  
   \rowcolor{gray!25}
  
  & HLPD                          & \textbf{0.9918} & \textbf{0.9955} & \textbf{0.9492} & 0.8480 & \textbf{0.9461} \\
  \midrule
\multirow{9}{*}{\textbf{PubMed}}
 
  & Likelihood    & 0.4204 & 0.6688 & 0.5956 & 0.6400 &0.5812 \\
  & Entropy       & 0.5752 & 0.4714 & 0.4644 & 0.4636 & 0.4936 \\
  & LogRank       & 0.3946 & 0.6480 & 0.5833
 & 0.6329 & 0.5647\\
  & LRR                & 0.3480 & 0.5611 & 0.5159 & 0.5875 & 0.5031 \\
  & NPR  $\diamondsuit$          & 0.3684 & 0.4047 & 0.4216& 0.4432& 0.4095 \\
  & DetectGPT  $\diamondsuit$  & 0.3259 & 0.5442 & 0.4466& 0.4941 & 0.4527 \\
  
  & Fast-DetectGPT     & 0.4240 & 0.7328
 & 0.6296 & 0.6934& 0.6200 \\

  & ImBD     & 0.7950 & 0.9152
 & \textbf{0.8165} & \textbf{0.7970}& \textbf{0.8309} \\
\rowcolor{gray!25}
  & HLPD                          &\textbf{0.9760} & \textbf{0.9689} & 0.7235 & 0.6349 & 0.8258 \\
  
  \midrule
\multirow{9}{*}{\textbf{Wikitext}}
  
  & Likelihood    & 0.5564 & 0.7704 & 0.6652 & 0.7508 & 0.6857 \\
  & Entropy       & 0.3647 & 0.2968& 0.3409 & 0.3025 & 0.3262 \\
  & LogRank       & 0.5381 & 0.7518 & 0.6439
 & 0.7376
 & 0.6679 \\
  & LRR                & 0.4779 & 0.6374 & 0.5547 & 0.6441 &0.5785 \\

  & NPR  $\diamondsuit$             & 0.4398 & 0.5980 & 0.5219 &0.5833& 	0.5358 \\
  & DetectGPT $\diamondsuit$  & 0.4515 & 0.6408& 0.5695 & 0.6000 & 0.5655 \\
  & Fast-DetectGPT    & 0.3683 &0.6988 &0.5620 & 0.6632
 & 0.5731 \\
   & ImBD     & 0.8074 & 0.9004
 & 0.8022 &0.8128 &  0.8307\\
  \rowcolor{gray!25}
 
  & HLPD                           &\textbf{0.9948} & \textbf{0.9874} & \textbf{0.9764} & \textbf{0.9327}& \textbf{0.9728} \\
  \rowcolor{gray!25}
 
\bottomrule
\end{tabular}
}
\caption{\textbf{Performance on \textit{open-source} model rewritten text.} Metric: AUROC. AUROC scores are averaged across the datasets generated by the polish task based on XSum, SQuAD, PubMed, Wikitext and WritingPrompts. The scoring model typically relies on Neo-2.7B as the source. In contrast, both NPR and DetectGPT leverage T5-3B  to create perturbations, while Fast-DetectGPT adopts GPT-J  as a surrogate model for generating samples.}
\label{tab:detals on open source model rewritten}
\end{table}

\begin{table}[h]
\scriptsize
\centering
\renewcommand{\arraystretch}{0.75} 

\begin{tabular}{llccccc}
\toprule
\rowcolor{white}
\multirow{2}{*}{\textbf{Model}} & \multirow{2}{*}{\textbf{Method}} &\multicolumn{4}{c}{\textbf{Task}}& \multirow{2}{*}{\textbf{Avg.}}\\
 & & \textbf{Rewrite} & \textbf{Polish} & \textbf{Expand} & \textbf{Generate}& \\
\midrule
\multirow{10}{*}{\textbf{GPT-3.5}} 
 & Likelihood    
   & 0.2774 & 0.4982 & 0.6105 & 0.9577 & 0.5860 \\
 & Entropy       
   & 0.6236 & 0.6742 & 0.5390 & 0.8867 & 0.6809 \\
 & LogRank       
   & 0.2528 & 0.4711 & 0.5849 & 0.9583 & 0.5668 \\
 & LRR                
   & 0.2158 & 0.4016 & 0.5039 & 0.9324 & 0.5134 \\
 & DNA-GPT         
   & 0.2720 & 0.5068 & 0.5370 & 0.9288 & 0.5612 \\
 & NPR             
   & 0.2873 & 0.5859 & 0.5856 & 0.9467 & 0.6014 \\
 & DetectGPT   
   & 0.3181 & 0.6135 & 0.5301 & 0.9203 & 0.5955 \\
 & Fast-DetectGPT 
   & 0.2683 & 0.7312 & 0.7801 & 0.9906 & 0.6920 \\
    & ImBD    
   & 0.8651 & 0.9849 & 0.7801 & 0.9900 &0.9600 \\
    \rowcolor{gray!25}
 & HLPD                       
   & \textbf{0.9998 } & \textbf{0.9923} & \textbf{0.9995} & \textbf{0.9991} & \textbf{0.9977} \\
\midrule
\multirow{10}{*}{\textbf{GPT-4o}}
 & Likelihood   
   & 0.4290 & 0.4396 & 0.5333 & 0.7585 & 0.5401 \\
 & Entropy      
   & 0.5351 & 0.6122 & 0.4867 & 0.4702 & 0.5261 \\
 & LogRank      
   & 0.4064 & 0.4002 & 0.5060 & 0.7486 & 0.5153 \\
 & LRR               
   & 0.3647 & 0.3095 & 0.4304 & 0.7070 & 0.4529 \\
 & DNA-GPT         
   & 0.4258 & 0.4974 & 0.5313 & 0.7528 & 0.5518 \\
 & NPR               
   & 0.4066 & 0.5065 & 0.5242 & 0.7304 & 0.5419 \\
 & DetectGPT   
   & 0.4361 & 0.6217 & 0.6318 & 0.7299 & 0.6053 \\
 & Fast-DetectGPT  
   & 0.3951 & 0.6293 & 0.6357 & 0.8282 & 0.6221 \\
    & ImBD   
   & 0.7995 & 0.9486 &0.9396 & 0.9988 &0.9216 \\
    \rowcolor{gray!25}
 & HLPD                         
   & \textbf{0.9619} & \textbf{0.9804} & \textbf{0.9316} & \textbf{0.9980} & \textbf{0.9680} \\
\midrule
\multirow{10}{*}{\textbf{Qwen2-7B}}
 & Likelihood   
   & 0.2741 & 0.2520 & 0.3404 & 0.4085 & 0.3188 \\
 & Entropy      
   & 0.6396 & 0.6726 & 0.3729 & 0.5053 & 0.5476 \\
 & LogRank      
   & 0.2564 & 0.2246 & 0.3179 & 0.7703 & 0.3928 \\
 & LRR                
   & 0.2376 & 0.1875 & 0.2396 & 0.7094 & 0.3435 \\
 & DNA-GPT          
   & 0.3253 & 0.3352 & 0.3558 & 0.7732 & 0.4474 \\
 & NPR                
   & 0.2443 & 0.3896 & 0.3705 & 0.7805 & 0.4462 \\
 & DetectGPT    
   & 0.2726 & 0.4983 & 0.4715 & 0.9756 & 0.5545 \\
 & Fast-DetectGPT    
   & 0.2853 & 0.5485 & 0.6000 & 0.9720 & 0.6015 \\
 
   & ImBD    
   &  0.8952 & 0.9589  &0.9720 & \textbf{1.0000} & 0.9565\\
   \rowcolor{gray!25}
 & HLPD                          
   & \textbf{0.9988} & \textbf{0.9951} & \textbf{0.9979} & \textbf{1.0000} & \textbf{0.9980} \\
\midrule
\multirow{10}{*}{\textbf{Llama-3-8B}}
 & Likelihood   
   & 0.5851 & 0.5685 & 0.6511 & 0.9496 & 0.6886 \\
 & Entropy      
   & 0.5165 & 0.6348 & 0.6030 & 0.9409 & 0.6738 \\
 & LogRank      
   & 0.5589 & 0.5412 & 0.6447 & 0.7499 & 0.6237 \\
 & LRR               
   & 0.4905 & 0.4530 & 0.5942 & 0.7291 & 0.5667 \\
 & DNA-GPT         
   & 0.5441 & 0.5599 & 0.6507 & 0.8928 & 0.6620 \\
 & NPR               
   & 0.4986 & 0.6144 & 0.6720 & 0.8700 & 0.6638 \\
 & DetectGPT   
   & 0.6356 & 0.6904 & 0.7632 & 0.9095 & 0.7497 \\
 & Fast-DetectGPT   
   & 0.6911 & 0.8192 & 0.9330 & 0.9828 & 0.8565 \\
   
    & ImBD    
   & 0.9710 & \textbf{0.9884} & \textbf{0.9821}&  \textbf{0.9989} &   \textbf{0.9851}\\
    \rowcolor{gray!25}
   & HLPD                         
   & \textbf{0.9966} & 0.9841 & 0.9117 & 0.9510 & 0.9359\\
\midrule
\multirow{10}{*}{\textbf{Mistral-7B}}
 & Likelihood   
   & 0.3613 & 0.4853 & 0.7056 & 0.9449 & 0.6243 \\
 & Entropy      
   & 0.6028 & 0.6539 & 0.4864 & 0.4905 & 0.5584 \\
 & LogRank      
   & 0.3399 & 0.3980 & 0.6282 & 0.7583 & 0.5311 \\
 & LRR                
   & 0.3071 & 0.3112 & 0.5095 & 0.9324 & 0.5150 \\
 & DNA-GPT          
   & 0.4006 & 0.4555 & 0.6705 & 0.9353 & 0.6155 \\
 & NPR                
   & 0.2888 & 0.4594 & 0.5858 & 0.9625 & 0.5741 \\
 & DetectGPT    
   & 0.3115 & 0.5480 & 0.6947 & 0.9753 & 0.6324 \\
 & Fast-DetectGPT    
   & 0.3938 & 0.7034 & 0.9161 & 0.9950 & 0.7529 \\
     & ImBD     
   &  0.8384 & 0.9671 &0.9946&  \textbf{1.0000} & 0.9500\\
   
    \rowcolor{gray!25}
 & HLPD                          
   & \textbf{0.9940} & \textbf{0.9947} & \textbf{0.9999} & 0.9989 & \textbf{0.9969} \\
\midrule
\multirow{10}{*}{\textbf{Deepseek-7B}}
 & Likelihood   
   & 0.5170 & 0.5438 & 0.7822 & 0.9788 & 0.7055 \\
 & Entropy      
   & 0.5862 & 0.6402 & 0.4609 & 0.8927 & 0.3950 \\
 & LogRank      
   & 0.5053 & 0.5288 & 0.7413 & 0.9670 & 0.6856 \\
 & LRR                
   & 0.4742 & 0.4680 & 0.6192 & 0.8250 & 0.5966 \\
 & DNA-GPT          
   & 0.4928 & 0.5837 & 0.7203 & 0.9462 & 0.6607 \\
 & NPR                
   & 0.4380 & 0.6177 & 0.7499 & 0.9746 & 0.6951 \\
 & DetectGPT    
   & 0.4512 & 0.6272 & 0.7429 & 0.9446 & 0.6915 \\
 & Fast-DetectGPT    
   & 0.6647 & 0.8177 & 0.8704 & 0.9906 & 0.8352 \\
     & ImBD     
&0.8739 &0.9764& \textbf{0.9766}& \textbf{1.0000}& 0.9567\\
    \rowcolor{gray!25}
 & HLPD                          
   & \textbf{0.9212}& \textbf{0.9902} & 0.9668 & 0.9755& \textbf{0.9634} \\
\bottomrule
\end{tabular}
\caption{\textbf{Detailed Results across diverse machine text revision tasks on \textit{XSum} dataset.} Metric: AUROC. The scoring model typically relies on Neo-2.7B  as the source. In contrast, both NPR and DetectGPT leverage T5-3B  to create perturbations, while Fast-DetectGPT adopts GPT-J  as a surrogate model for generating samples.
}
\label{tab:performance across different Models and Tasks.}
\end{table}

\end{document}